\documentclass[10pt,journal,compsoc]{IEEEtran}

\usepackage{times}
\usepackage{epsfig}
\usepackage{graphicx}
\usepackage{amsmath}
\usepackage{amssymb}
\usepackage{subfigure}
\usepackage{algorithmic}
\usepackage{algorithm}
\usepackage{url}
\usepackage{multirow}
\usepackage{booktabs}
\usepackage{color}
\usepackage{ragged2e}

\usepackage{xspace}
\makeatletter
\DeclareRobustCommand\onedot{\futurelet\@let@token\@onedot}
\def\@onedot{\ifx\@let@token.\else.\null\fi\xspace}
\def\eg{{e.g}\onedot} 
\def\ie{{i.e}\onedot}

\def\etal{{et al}\onedot}
\def \x {\mathbf{x}}
\def\M{\mathcal M}
\makeatother

\begin{document}

\title{ArcFace: Additive Angular Margin Loss for Deep Face Recognition}

\author{
Jiankang~Deng,
Jia~Guo,
Jing~Yang,
Niannan~Xue,
Irene~Kotsia,
and~Stefanos~Zafeiriou
\IEEEcompsocitemizethanks{\IEEEcompsocthanksitem
Corresponding author: Jiankang Deng, E-mail: j.deng16@imperial.ac.uk \protect\\
J. Deng, N. Xue and S. Zafeiriou are with the Department of Computing, Imperial College London, UK. J. Guo is with InsightFace. \textcolor{black}{J. Yang is with Department of Computer Science, University of Nottingham. I. Kotsia is with Cogitat.}
}
\thanks{Manuscript received on November 28, 2020; revised on May 4, 2021; accepted on June 4, 2021.}
}

\markboth{Journal of \LaTeX\ Class Files,~Vol.~14, No.~8, August~2015}%
{Shell \MakeLowercase{\textit{et al.}}: Bare Demo of IEEEtran.cls for Computer Society Journals}

\IEEEtitleabstractindextext{
\begin{abstract}
\justifying  
Recently, a popular line of research in face recognition is adopting margins in the well-established softmax loss function to maximize class separability. In this paper, we first introduce an Additive Angular Margin Loss (ArcFace), which not only has a clear geometric interpretation but also significantly enhances the discriminative power. Since ArcFace is susceptible to the massive label noise, we further propose sub-center ArcFace, in which each class contains $K$ sub-centers and training samples only need to be close to any of the $K$ positive sub-centers. Sub-center ArcFace encourages one dominant sub-class that contains the majority of clean faces and non-dominant sub-classes that include hard or noisy faces. Based on this self-propelled isolation, we boost the performance through automatically purifying raw web faces under massive real-world noise. Besides discriminative feature embedding, we also explore the inverse problem, mapping feature vectors to face images. Without training any additional generator or discriminator, the pre-trained ArcFace model can generate identity-preserved face images for both subjects inside and outside the training data only by using the network gradient and Batch Normalization (BN) priors. Extensive experiments demonstrate that ArcFace can enhance the discriminative feature embedding as well as strengthen the generative face synthesis.
\end{abstract}
\begin{IEEEkeywords}
Large-scale Face Recognition, Additive Angular Margin, Noisy Labels, Sub-class, Model Inversion
\end{IEEEkeywords}}
\maketitle
\IEEEdisplaynontitleabstractindextext
\IEEEpeerreviewmaketitle

\section{Introduction}

\IEEEPARstart{F}{ace} representation using DCNN embedding is the method of choice for face recognition \cite{sun2014deep,taigman2014deepface,schroff2015facenet,parkhi2015deep,masi2018deep,guo2019survey}. DCNNs map the face image, typically after a pose normalization step \cite{zhang2016joint,deng2020retinaface}, into a feature that should have small intra-class and large inter-class distance. There are two main lines of research to train DCNNs for face recognition. Some train a multi-class classifier which can separate different identities in the training set, such by using a softmax classifier \cite{taigman2014deepface,parkhi2015deep,cao2017vggface2,masi2016we,masi2019face}, and the others learn directly an embedding, such as the triplet loss \cite{schroff2015facenet}. Based on the large-scale training data and the elaborate DCNN architectures, both the softmax-loss-based methods \cite{cao2017vggface2} and the triplet-loss-based methods \cite{schroff2015facenet} can obtain excellent performance on face recognition. However, both the softmax loss and the triplet loss have some drawbacks. For the softmax loss: (1) the learned features are separable for the closed-set classification problem but not discriminative enough for the open-set face recognition problem; (2) the size of the linear transformation matrix $W \in \mathbb{R}^{d \times N}$ increases linearly with the identities number $N$. For the triplet loss: (1) there is a combinatorial explosion in the number of face triplets especially for large-scale datasets, leading to a significant increase in the number of iteration steps; (2) semi-hard sample mining is a quite difficult problem for effective model training.

\begin{figure*}[!t]
\centering
\includegraphics[width=0.9\linewidth]{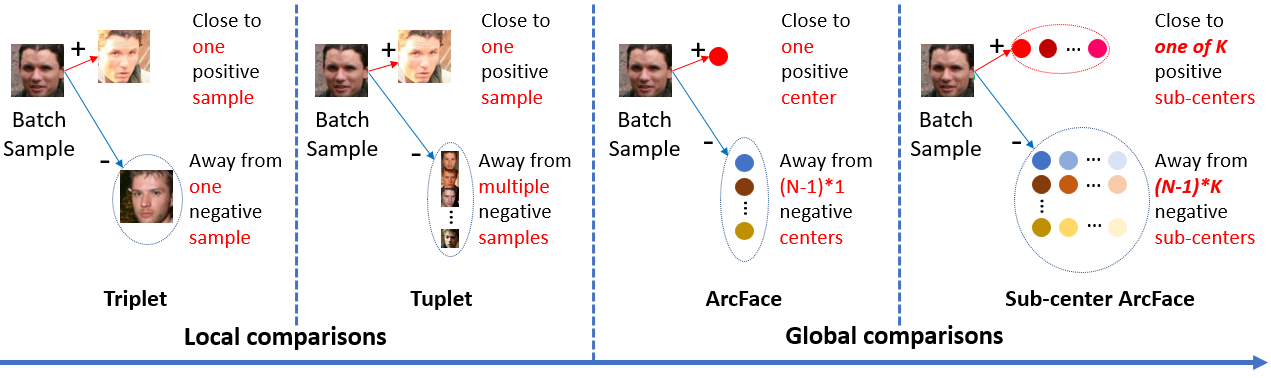}
\caption{Comparisons of Triplet \cite{schroff2015facenet}, Tuplet \cite{sohn2016improved}, ArcFace and sub-center ArcFace. Triplet and Tuplet conduct local sample-to-sample comparisons with Euclidean margins within the mini-batch. By contrast, ArcFace and sub-center ArcFace conduct global sample-to-class and sample-to-subclass comparisons with angular margins.}
\vspace{-4mm}
\label{fig:comparisons}
\end{figure*}

To adopt margin benefit but avoid the sampling problem in the Triplet loss~\cite{schroff2015facenet}, recent methods~\cite{liu2017sphereface,tencent2017CosineFace,wang2018additive} focus on incorporating margin penalty into a more feasible framework, the softmax loss, which has global sample-to-class comparisons within the multiplication step between the embedding feature and the linear transformation matrix. Naturally, each column of the linear transformation matrix is viewed as a class center representing a certain class. Sphereface \cite{liu2017sphereface} introduces the important idea of angular margin, however their loss function requires a series of approximations, which results in an unstable training of the network. In order to stabilize training, they propose a hybrid loss function which includes the standard softmax loss. Empirically, the softmax loss dominates the training process, because the integer-based multiplicative angular margin makes the target logit curve very precipitous and thus hinders convergence.

In this paper, we propose an Additive Angular Margin loss~ \textcolor{black}{\cite{deng2019arcface}} to stabilize the training process and further improve the discriminative power of the face recognition model. More specifically, the dot product between the DCNN feature and the last fully connected layer is equal to the cosine distance after feature and center normalization. We utilize the arc-cosine function to calculate the angle between the current feature and the target center. Afterwards, we \textcolor{black}{introduce} an additive angular margin to the target angle, and we get the target logit back again by the cosine function. Then, we re-scale all logits by a fixed feature norm, and the subsequent steps are exactly the same as in the softmax loss. Due to the exact correspondence between the angle and arc in the normalized hypersphere, our method can directly optimize the geodesic distance margin, thus we call it ArcFace. 

Even though impressive performance has been achieved by the margin-based softmax methods~\cite{deng2017marginal,liu2017sphereface,tencent2017CosineFace,wang2018additive}, they all need to be trained on well-annotated clean datasets~\cite{wang2018devil}, which require intensive human efforts.
Wang~\etal~\cite{wang2018devil} found that faces with label noise significantly degenerate the recognition accuracy and manually built a high-quality dataset including 1.7M images of 59K celebrities. However, it took 50 annotators to work continuously for one month to clean the dataset, which further demonstrates the difficulty of obtaining a large-scale clean dataset for face recognition. Since accurate manual annotations can be expensive~\cite{wang2018devil}, learning with massive noisy data has recently attracted much attention~\cite{hu2019noise,Zhong2019Unequal,wang2019co}. However, computing time-varying weights for samples~\cite{hu2019noise} or designing piece-wise loss functions~\cite{Zhong2019Unequal} according to the current model's predictions can only alleviate the influence from noisy data to some extent as the robustness and improvement depend on the initial performance of the model. Besides, the co-mining method~\cite{wang2019co} requires to train twin networks together thus it is less practical for training large models on large-scale datasets. 

To improve the robustness under massive real-world noise, we relax the intra-class constraint of forcing all samples close to the corresponding positive centers by introducing sub-classes into ArcFace~\textcolor{black}{\cite{deng2020sub}}. \textcolor{black}{As illustrated in Figure \ref{fig:comparisons},} we design $K$ sub-centers for each class and the training sample only needs to be close to any of the $K$ positive sub-centers instead of the only one positive center. If a training face is a noisy sample, it does not belong to the corresponding positive class. In ArcFace, this noisy sample generates a large wrong loss value, which impairs the model training. In sub-center ArcFace, the intra-class constraint enforces the training sample to be close to one of the multiple positive sub-centers but not all of them. The noise is likely to form a non-dominant sub-class and will not be enforced into the dominant sub-class. Therefore, sub-center ArcFace is more robust to noise. In our experiments, we find the proposed sub-center ArcFace can encourage one dominant sub-class that contains the majority clean faces and multiple non-dominant sub-classes that include hard or noisy faces. This automatic isolation can be directly employed to clean the training data through dropping non-dominant sub-centers and high-confident noisy samples. Based on the proposed sub-center ArcFace, we can automatically obtain large-scale clean training data from raw web face images to further improve the discriminative power of the face recognition model.

\textcolor{black}{In Figure \ref{fig:comparisons}, we compare the differences between Triplet \cite{schroff2015facenet}, Tuplet \cite{sohn2016improved}, ArcFace and sub-center ArcFace. Triplet loss \cite{schroff2015facenet} only considers local sample-to-sample comparisons with Euclidean margins within the mini-batch. Tuplet loss \cite{sohn2016improved} further enhances the comparisons by using all of the negative pairs within the mini-batch. By contrast, the proposed ArcFace and sub-center ArcFace conduct global sample-to-class and sample-to-subclass comparisons with angular margins.}

As the proposed ArcFace is effective for the mapping from the face image to the discriminative feature embedding, we are also interested in the inverse problem: the mapping from a low-dimensional latent space to a highly nonlinear face space. 
Synthesizing face images \cite{cole2017synthesizing,nhan2015beyond,duong2019deep,dosovitskiy2016inverting,zhmoginov2016inverting,mai2018reconstruction,duong2020vec2face} has recently brought much attention from the community.
DeepDream \cite{mordvintsev2015inceptionism} is proposed to transform a random input to yield a high
output activation for a chosen class by employing the gradient from the pre-trained classification model and some regularizers (\eg total variance \cite{mahendran2015understanding} for maintaining piece-wise constant patches). Even though DeepDream can keep the selected output response high to preserve identity, the resulting faces are not realistic, lacking natural face statistics. \textcolor{black}{Inspired by the pioneer generative face recognition model (Eigenface~\cite{turk1991face}) and recent data-free methods \cite{yin2020dreaming,haroush2020knowledge,cai2020zeroq} for restoring ImageNet images, we employ the statistic prior (\eg mean and variance stored in the BN layers) to constrain the face generation.} In this paper, we show that the proposed ArcFace can also enhance the generative power. Without training any additional generator or discriminator like in Generative Adversarial Networks (GANs) \cite{goodfellow2014generative}, the pre-trained ArcFace model can generate identity-preserved and \textcolor{black}{visually reasonable} face images only by using the gradient and BN priors.

The advantages of the proposed methods can be summarized as follows:

\noindent{\bf \textcolor{black}{Intuitive}.} ArcFace directly optimizes the geodesic distance margin by virtue of the exact correspondence between the angle and arc in the normalized hypersphere. The proposed additive angular margin loss can intuitively enhance the intra-class compactness and inter-class discrepancy during discriminative learning of face feature embedding.  

\noindent{\bf Economical.} We introduce sub-class into ArcFace to improve its robustness under massive real-world noises. The proposed sub-center ArcFace can automatically clean the large-scale raw web faces (\eg MS1MV0 \cite{guo2016ms} and Celeb500K~\cite{cao2018celeb}) without expensive and intensive human efforts.
The automatically cleaned training data, named IBUG-500K, has been released to facilitate future research.

\noindent{\bf Easy.} ArcFace only needs several lines of code and is extremely easy to implement in the computational-graph-based deep learning frameworks, \eg MxNet \cite{chen2015mxnet}, Pytorch \cite{paszke2017automatic} and Tensorflow \cite{abadi2016tensorflow}. Furthermore, contrary to the works in \cite{liu2017sphereface,liu2016large}, ArcFace does not need to be combined with other loss functions in order to have stable convergence.

\noindent{\bf Efficient.} ArcFace only adds negligible computational complexity during training. The proposed center parallel strategy can easily support millions of identities for training on a single server (8 GPUs).

\noindent{\bf Effective.} Using IBUG-500K as the training data, ArcFace achieves state-of-the-art performance on ten face recognition benchmarks including large-scale image and video datasets collected by us. Impressively, our model reaches $97.27\%$ TPR@FPR=1e-4 on IJB-C. Code and pre-trained models have been made available.

\noindent{\bf Engaging.} ArcFace can not only enhance the discriminative power but also strengthen the generative power. By accessing the network gradient and employing the statistic priors stored in the BN layers, the pre-trained ArcFace model can restore identity-preserved and \textcolor{black}{visually plausible} face images for both subjects inside and outside the training data.

\section{Related Work}

\noindent{\bf Face Recognition with Margin Penalty.} \textcolor{black}{As shown in Figure \ref{fig:comparisons},} the pioneering work \cite{schroff2015facenet} uses the Triplet loss to exploit triplet data such that faces from the same class are closer than faces from different classes by a clear Euclidean distance margin. Even though the Triplet loss makes perfect sense for face recognition, the sample-to-sample comparisons are local within mini-batch and the training procedure for the Triplet loss is very challenging as there is a combinatorial explosion in the number of triplets especially for large-scale datasets, requiring effective sampling strategies to select informative mini-batch~\cite{rippel2015metric,schroff2015facenet} and choose representative triplets within the mini-batch~\cite{oh2016deep,sohn2016improved}.
\textcolor{black}{
As the Triplet loss trained with semi-hard negative mining converges slower due to the ignorance of too many examples, a double-margin contrastive loss is proposed in \cite{wu2017sampling} to explore more informative and stable examples by distance weighted sampling, thus it converges faster and more accurately.}
Some other works tried to reduce the total number of triplets with proxies~\cite{movshovitz2017no,qian2018large}, \ie, sample-to-sample comparison is changed into sample-to-proxy comparison. However, sampling and proxy methods only optimize the embedding of partial classes instead of all classes in one iteration step.

Margin-based softmax methods~\cite{liu2017sphereface,deng2017marginal,tencent2017CosineFace,wang2018additive} focused on incorporating margin penalty into a more feasible framework, softmax loss, which has extensive sample-to-class comparisons. Compared to deep metric learning methods (\eg, Triplet~\cite{schroff2015facenet}, Tuplet~\cite{oh2016deep,sohn2016improved}), margin-based softmax methods conduct global comparisons at the cost of memory consumption on holding the center of each class \textcolor{black}{as illustrated in Figure \ref{fig:comparisons}.} Sample-to-class comparison is more efficient and stable than sample-to-sample comparison as (1) the class number is much smaller than sample number, and (2) each class can be represented by a smoothed center vector which can be updated online during training. To further improve the margin-based softmax loss, recent works focus on the exploration of adaptive parameters \cite{zhang2019p2sgrad,zhang2019adacos,liu2019adaptiveface}, inter-class regularization \cite{zhao2019regularface,duan2019uniformface}, mining \cite{wang2019mis,huang2020curricularface}, grouping \cite{kim2020groupface}, etc.

\noindent{\bf Face Recognition under Noise.} Most of the face recognition datasets~\cite{yi2014learning,guo2016ms,cao2017vggface2,cao2018celeb} are downloaded from the Internet by searching a pre-defined celebrity list, and the original labels are likely to be ambiguous and inaccurate~\cite{wang2018devil}.
Learning with massive noisy data has recently drawn much attention in face recognition~\cite{wu2018light,hu2019noise,Zhong2019Unequal,wang2019co} as accurate manual annotations can be expensive~\cite{wang2018devil} or even unavailable.

Wu~\etal~\cite{wu2018light} proposed a semantic bootstrap strategy, which re-labels the noisy samples according to the probabilities of the softmax function. However, automatic cleaning by the bootstrapping rule requires time-consuming iterations (\eg twice refinement steps are used in~\cite{wu2018light}) and the labelling quality is affected by the capacity of the original model. Hu~\etal~\cite{hu2019noise} found that the cleanness possibility of a sample can be dynamically reflected by its position in the target logit distribution and presented a noise-tolerant end-to-end paradigm by employing the idea of weighting training samples. Zhong~\etal~\cite{Zhong2019Unequal} devised a noise-resistant loss by introducing a hypothetical training label, which is a convex combination of the original label with probability $\rho$ and the predicted label by the current model with probability $1-\rho$.
However, computing time-varying fusion weight~\cite{hu2019noise} and designing piece-wise loss~\cite{Zhong2019Unequal} contain many hand-designed hyper-parameters. Besides, re-weighting methods are susceptible to the performance of the initial model. Wang~\etal~\cite{wang2019co} proposed a co-mining strategy which uses the loss values as the cue to simultaneously detect noisy labels, exchange the high-confidence clean faces to alleviate the error accumulation caused by the sampling bias, and re-weight the predicted clean faces to make them dominate the discriminative model training. However, the co-mining method requires training twin networks simultaneously and it is challenging to train large networks (\eg ResNet100~\cite{he2016deep}) on a large-scale dataset (\eg MS1MV0~\cite{guo2016ms} and Celeb500K~\cite{cao2018celeb}). 

\noindent{\bf Face Recognition with Sub-classes.} Practices and theories that lead to ``sub-class''  have been studied for a long
time~\cite{zhu2004optimal,zhu2006subclass}. The concept of ``sub-class'' applied in face recognition was first introduced in~\cite{zhu2004optimal,zhu2006subclass}, 
where a mixture of Gaussians was used to approximate the underlying distribution of each class. For instance, a person’s face images may be frontal view or side view, resulting in different modalities when all images are represented in the same data space. In~\cite{zhu2004optimal,zhu2006subclass}, experimental results showed that subclass divisions can be used to effectively adapt to different face modalities thus improve the performance of face recognition.
Wan~\etal~\cite{wan2017separability} further proposed a separability criterion to divide every class into sub-classes, which have much less overlaps. The new within-class scatter can represent multi-modality information, therefore optimizing this within-class scatter will separate different modalities more clearly and further increase the accuracy of face recognition. However, these work~\cite{zhu2004optimal,zhu2006subclass,wan2017separability} only employed hand-designed feature descriptor on tiny under-controlled datasets. 

Concurrent with our work, Softtriple~\cite{qian2019softtriple} presents a multi-center softmax loss with class-wise regularizer.
These multi-centers can depict the hidden distribution of the data~\cite{muller2020subclass} due to the fact that they can capture the complex geometry of the original data and help reduce the intra-class variance. On the fine-grained visual retrieval problem, the Softtriple~\cite{qian2019softtriple} loss achieves better performance than the softmax loss as capturing local clusters is essential for this task.
Even though the concept of ``sub-class'' has been employed in face recognition~\cite{zhu2004optimal,zhu2006subclass,wan2017separability} and fine-grained visual retrieval~\cite{qian2019softtriple}, none of these work has considered the large-scale (\eg 0.5 million classes) face recognition problem under massive noise (\eg around $50\%$ noisy samples within the training data). 

\begin{figure*}[!t]
\centering
\includegraphics[width=0.9\linewidth]{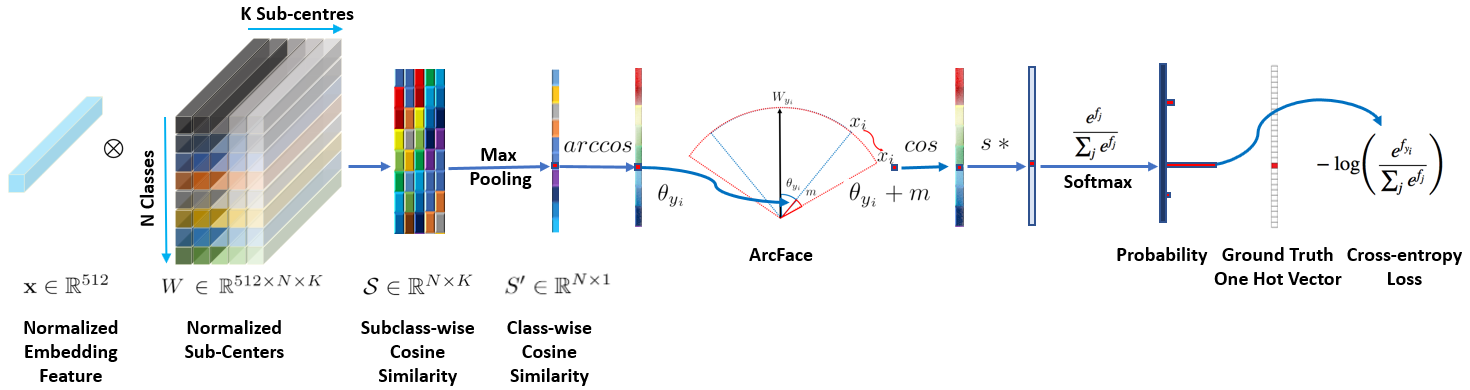}
\caption{Training the deep face recognition model by the proposed ArcFace loss ($K$=1) and sub-center ArcFace loss (\eg $K$=3). Based on a $\ell_2$ normalization step on both embedding feature $\x_i \in \mathbb{R}^{512}$ and all sub-centers $W \in \mathbb{R}^{512 \times N \times K}$, we get the subclass-wise similarity score $\mathcal{S} \in \mathbb{R}^{N \times K}$ by a matrix multiplication $W^T \x_i$. After a max pooling step, we can easily get the class-wise similarity score $\mathcal{S'} \in \mathbb{R}^{N \times 1}$. Afterwards, we calculate the $arccos\theta_{y_i}$ and get the angle between the feature $x_i$ and the ground truth center $W_{y_i}$. Then, we add an angular margin penalty $m$ on the target (ground truth) angle $\theta_{y_i}$. After that, we calculate $\cos(\theta_{y_i}+m)$ and multiply all logits by the feature scale $s$. Finally, the logits go through the softmax function and contribute to the cross entropy loss.}
\vspace{-4mm}
\label{fig:subcenterarcface}
\end{figure*}

\noindent{\bf Face Synthesis by Model Inversion.} \textcolor{black}{Identity-preserving face generation \cite{shen2018faceid,bao2018towards,li2020advancing,duong2020vec2face} has been extensively explored under the framework of GAN \cite{goodfellow2014generative}. Even though GAN models can yield high-fidelity images \cite{brock2018large,karras2019style}, training a GAN's generator requires access to the original data. Due to the emerging concern of data privacy, an alternative line of work in security focuses on model inversion, that is, image synthesis from a single CNN. 
Model inversion can not only help researchers to visualize neural networks to understand their properties \cite{samek2019explainable} but also can be used for data-free distillation, quantization and pruning \cite{yin2020dreaming,haroush2020knowledge,cai2020zeroq}.}
Fredrikson \etal \cite{fredrikson2015model} propose the model inversion attack to obtain class images from a network through a gradient descent on the input. As the pixel space is so large compared to the feature space, optimizing the image pixels by gradient descent \cite{mahendran2015understanding} requires heavy regularization terms, such as total variation \cite{mahendran2015understanding} or Gaussian blur \cite{yosinski2015understanding}. Even though previous model inversion methods \cite{fredrikson2015model,mordvintsev2015inceptionism} can transform an input image (random noise or a natural image) to yield a high output activation for a chosen class, it leaves intermediate representations constraint-free. Therefore, the resulting images are not realistic, lacking natural image statistics.

The pioneer generative face recognition model is Eigenface~\cite{turk1991face}, which can project a training face image or a new face image (mean-subtracted) on the eigenfaces and thereby record how that face differs from the mean face. The eigenvalue associated with each eigenface represents how much the image vary from the mean image in that direction. The recognition process with the eigenface method is to project query images into the face-space spanned by eigenfaces calculated, and to find the closest match to a face class in that face-space. Even though raw pixel features used in Eigenface are substituted by the deep convolutional features, the procedure of employing the statistic prior (\eg mean and variance) to reconstruct face images can be an inspiration. \textcolor{black}{Recently, \cite{yin2020dreaming,haroush2020knowledge,cai2020zeroq} have proposed a data-free method employing the statistics (\eg mean and variance) stored in the BN layers to restore ImageNet images.} 
Inspired by these works, we synthesize face images by inverting the pre-trained ArcFace model and considering the face prior (\eg mean and variance) stored in the BN layers. 

\section{Proposed Approach}

\subsection{ArcFace}

The most widely used classification loss function, softmax loss, is presented as follows:
\begin{equation}
L_1=-\log\frac{e^{W^T_{y_i} x_i+b_{y_i}}}{\sum_{j=1}^{N}e^{W^T_j x_i+b_j}},
\label{eq:softmax}
\vspace{-1mm}
\end{equation}
where $x_i\in\mathbb{R}^d$ denotes the deep feature of the $i$-th sample, belonging to the $y_i$-th class. The embedding feature dimension $d$ is set to $512$ in this paper following \cite{wen2016discriminative,zhang2016range,liu2017sphereface,tencent2017CosineFace}. $W_j\in\mathbb{R}^d$ denotes the $j$-th column of the weight $W \in \mathbb{R}^{d \times N}$, $b_j\in\mathbb{R}^N $ is the bias term, and the class number is $N$. Traditional softmax loss is widely used in deep face recognition \cite{parkhi2015deep,cao2017vggface2}. However, the softmax loss function does not explicitly optimize the feature embedding to enforce higher similarity for intra-class samples and diversity for inter-class samples, which results in a performance degeneration for deep face recognition under large intra-class appearance variations (\eg pose variations \cite{sengupta2016frontal,zheng2018cross} and age gaps \cite{Moschoglou2017AgeDB,zheng2017cross}) and large-scale test scenarios \cite{kemelmacher2016megaface,whitelam2017iarpa,maze2018iarpa}.

For simplicity, we fix the bias $b_j=0$ as in \cite{liu2017sphereface}. Then, we transform the logit \cite{pereyra2017regularizing} as $W^T_j x_i=\left \| W_j \right \|\left \| x_i \right \|\cos\theta_j$, where $\theta_j$ is the angle between the weight $W_j$ and the feature $x_i$. Following \cite{liu2017sphereface,tencent2017CosineFace,wang2017normface}, we fix the individual weight $\left \| W_j \right \|=1$ by $\ell_2$ normalization. Following \cite{ranjan2017l2,tencent2017CosineFace,wang2017normface,wang2018additive}, we also fix the embedding feature $\left \| x_i \right \|$ by $\ell_2$ normalization and re-scale it to $s$. 
The normalization step on features and weights makes the predictions only depend on the angle between the feature and the weight. The learned embedding features are thus distributed on a hypersphere with a radius of $s$.
\begin{equation}
{L_2}=-\log\frac{e^{s\cos\theta_{y_i}}}{e^{s\cos\theta_{y_i}}+\sum_{j=1,j\neq  y_i}^{N}e^{s\cos\theta_{j}}}.
\label{eq:l2featweight}
\vspace{-1mm}
\end{equation}

Since the embedding features are distributed around each feature center on the hypersphere, we \textcolor{black}{employ} an additive angular margin penalty $m$ between $x_i$ and $W_{y_i}$ to simultaneously enhance the intra-class compactness and inter-class discrepancy as illustrated in Figure~\ref{fig:subcenterarcface}. Since the proposed additive angular margin penalty is equal to the geodesic distance margin penalty in the normalized hypersphere, we name our method as ArcFace. 
\begin{equation}
{L_3}=-\log\frac{e^{s\cos(\theta_{y_i}+m)}}{e^{s\cos(\theta_{y_i}+m)}+\sum_{j=1,j\neq  y_i}^{N}e^{s\cos\theta_{j}}}.
\label{eq:arcface}
\vspace{-1mm}
\end{equation}

\begin{figure}
\centering
\subfigure[\textcolor{black}{Norm-Softmax}]{
\label{fig:compactnesssoftmaxnorm}
\includegraphics[width=0.2\textwidth]{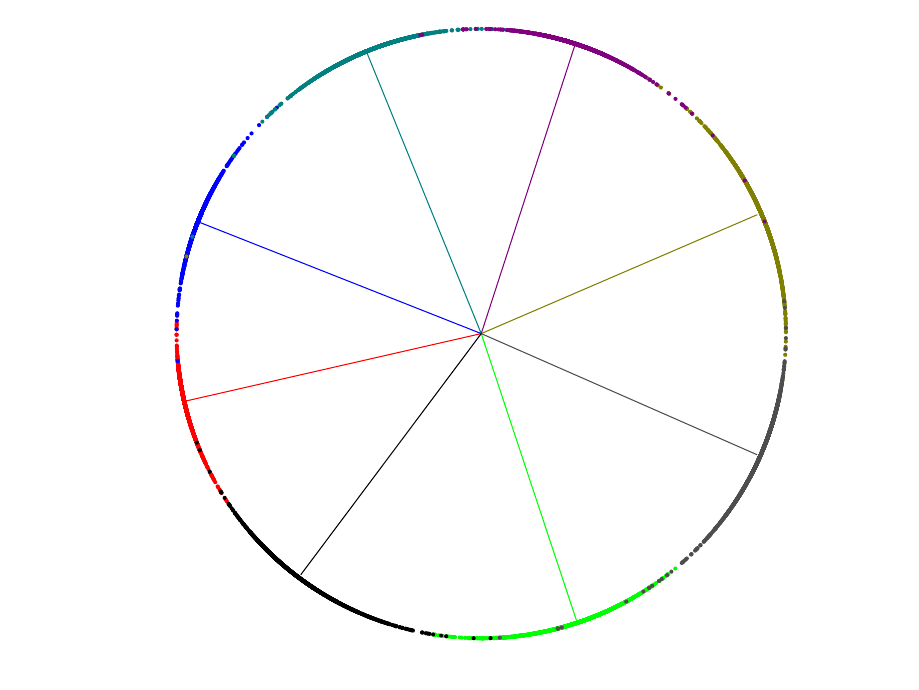}}
\subfigure[ArcFace]{
\label{fig:compactnessarcfacenorm}
\includegraphics[width=0.2\textwidth]{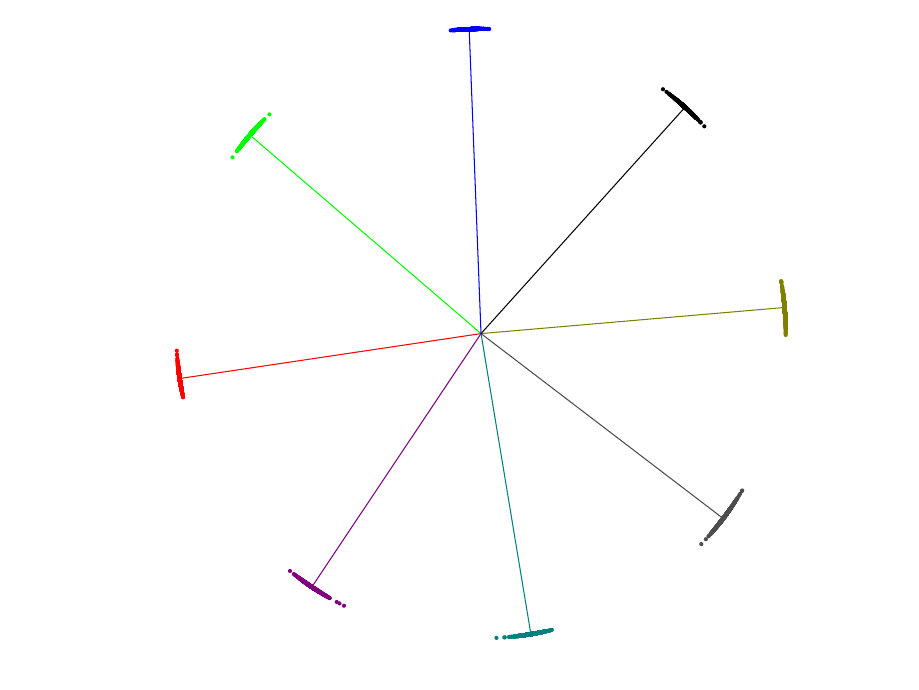}}
\caption{Toy examples under the \textcolor{black}{Norm-Softmax} and ArcFace loss on 8 identities with 2D features. Dots indicate samples and lines refer to the center direction of each identity. Based on the feature normalization, all face features are pushed to the arc space with a fixed radius. The geodesic distance margin between closest classes becomes evident as the additive angular margin penalty is incorporated.}
\vspace{-4mm}
\label{fig:compactness}
\end{figure}

We select face images from 8 different identities containing enough samples (around 1,500 images/class) to train 2-D feature embedding networks with the \textcolor{black}{Norm-Softmax} and ArcFace loss, respectively. As illustrated in Figure \ref{fig:compactness}, all face features are pushed to the arc space with a fixed radius based on the feature normalization. The \textcolor{black}{Norm-Softmax} loss provides roughly separable feature embedding but produces noticeable ambiguity in decision boundaries, while the proposed ArcFace loss can obviously enforce a more evident margin between the nearest classes.

\begin{figure}
\centering
\subfigure[$\theta_j$ Distributions]{
\label{fig:thetadistribution}
\includegraphics[width=0.23\textwidth]{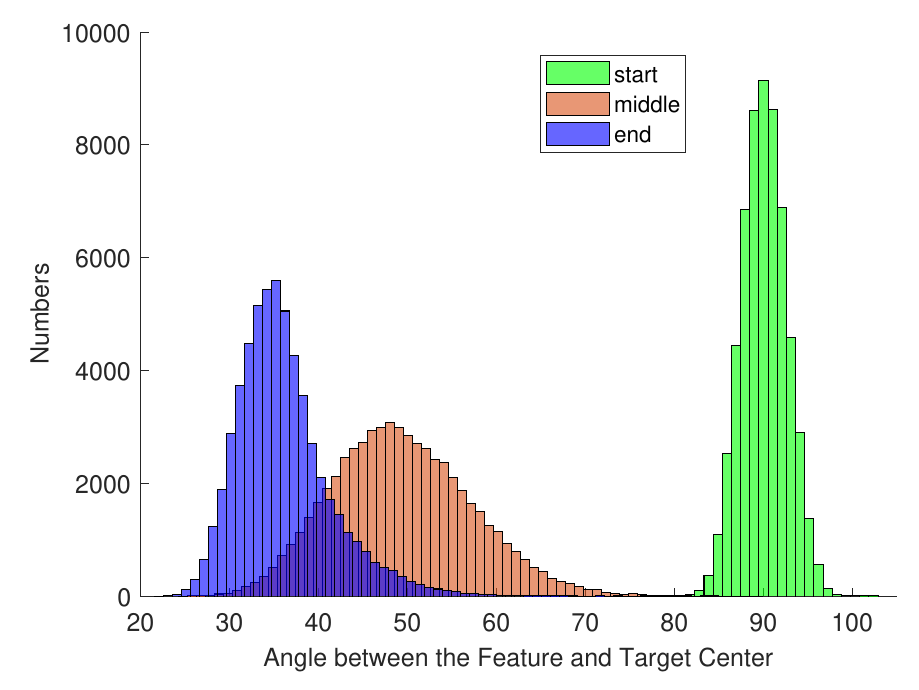}}
\subfigure[Target Logits Curves]{
\label{fig:soatargetlogit}
\includegraphics[width=0.23\textwidth]{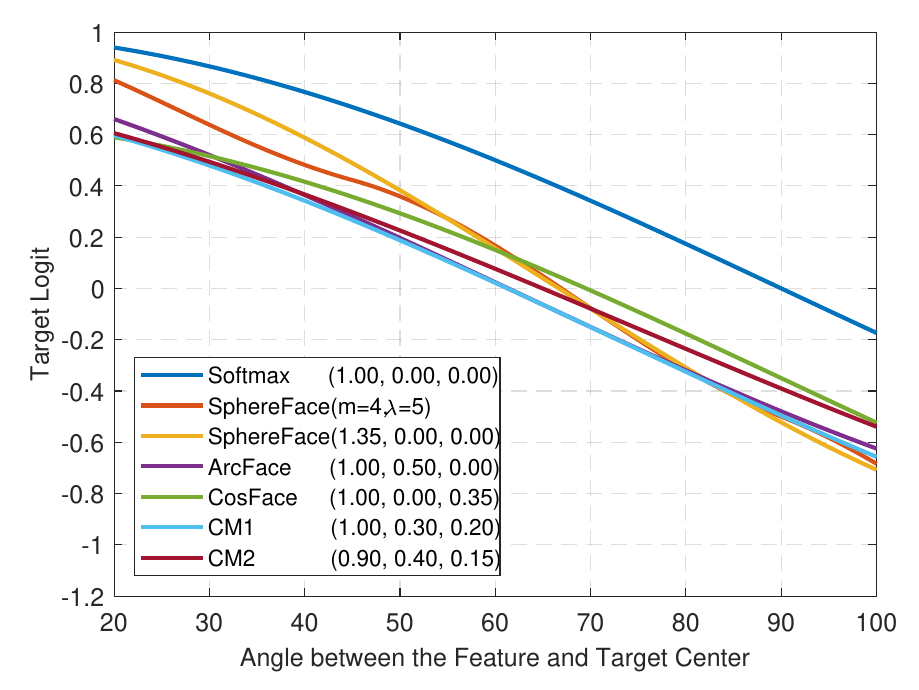}}
\caption{Target logit analysis. (a) $\theta_j$ distributions from start to end during ArcFace training. (2) Target logit curves for softmax, SphereFace, ArcFace, CosFace and combined margin penalty ($\cos(m_1 \theta+m_2)-m_3$).}
\vspace{-4mm}
\label{fig:targetlogitscurves}
\end{figure}

\begin{figure}[t!]
\centering
\includegraphics[width=0.9\linewidth]{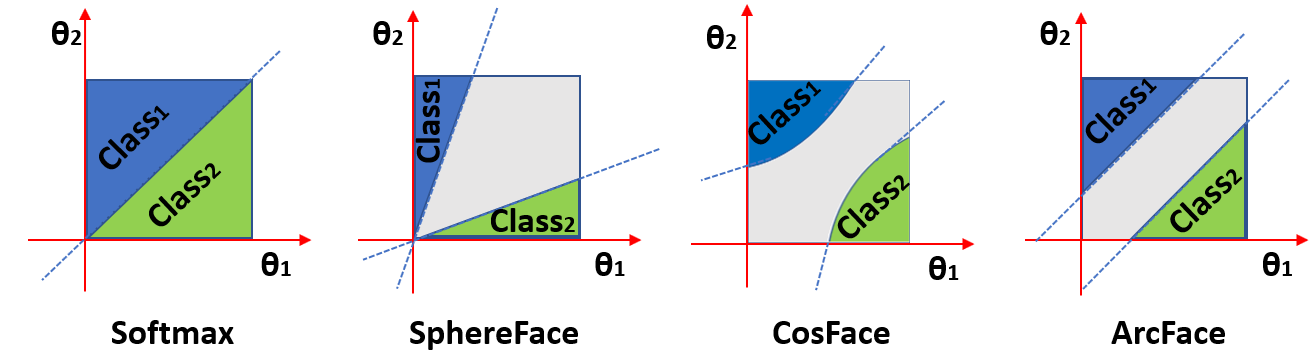}
\caption{Decision margins of different loss functions under binary classification case. The dashed line represents the decision boundary, and the grey areas are the decision margins.}
\vspace{-4mm}
\label{fig:binarymargin}
\end{figure}

\noindent{\bf Numerical Similarity.} In SphereFace \cite{liu2017sphereface,liu2016large}, ArcFace, and CosFace \cite{tencent2017CosineFace,wang2018additive}, three different kinds of margin penalty are proposed, \eg multiplicative angular margin $m_1$, additive angular margin $m_2$, and additive cosine margin $m_3$, respectively. From the view of numerical analysis, different margin penalties, no matter add on the angle \cite{liu2017sphereface} or cosine space \cite{tencent2017CosineFace}, all enforce the intra-class compactness and inter-class diversity by penalizing the target logit \cite{pereyra2017regularizing}. In Figure \ref{fig:soatargetlogit}, we plot the target logit curves of SphereFace, ArcFace and CosFace under their best margin settings. We only show these target logit curves within $[20^{\circ}, 100^{\circ}]$ because the angles between $W_{y_i}$ and $x_i$ start from around $90^{\circ}$ (random initialization) and end at around $30^{\circ}$ during ArcFace training as shown in Figure \ref{fig:thetadistribution}. Intuitively, there are three 
numerical factors in the target logit curves that affect the performance, \ie the starting point, the end point and the slope. 

By combining all of the margin penalties, we implement SphereFace, ArcFace and CosFace in a united framework with $m_1$, $m_2$ and $m_3$ as the hyper-parameters. 
\begin{equation}
{L_4}=-\log\frac{e^{s(\cos(m_1\theta_{y_i}+m_2)-m_3)}}{e^{s(\cos(m_1\theta_{y_i}+m_2)-m_3)}+\sum_{j=1,j\neq  y_i}^{N}e^{s\cos\theta_{j}}}.
\label{eq:combineloss}
\vspace{-1mm}
\end{equation}
As shown in Figure \ref{fig:soatargetlogit}, by combining all of the above-motioned margins ($\cos(m_1 \theta+m_2)-m_3$), we can easily get some other target logit curves which also achieve high performance. 

\noindent{\bf Geometric Difference.} Despite the numerical similarity between ArcFace and previous works, the proposed additive angular margin has a better geometric attribute as the angular margin has the exact correspondence to the geodesic distance. As illustrated in Figure \ref{fig:binarymargin}, we compare the decision boundaries under the binary classification case. The proposed ArcFace has a constant linear angular margin throughout the whole interval. By contrast, SphereFace and CosFace only have a nonlinear angular margin. 

The minor difference in margin designs can have a significant influence on model training. For example, the original SphereFace \cite{liu2017sphereface} employs an annealing optimization strategy. To avoid divergence at the beginning of training, joint supervision from softmax is used in SphereFace to weaken the multiplicative integer margin penalty. We implement a new version of SphereFace without the integer requirement on the margin by employing the arc-cosine function instead of using the complex double angle formula. In our implementation, we find that $m=1.35$ can obtain similar performance compared to the original SphereFace without any convergence difficulty. 

\noindent{\bf Other Intra and Inter Losses.} Other loss functions can be designed based on the angular representation of features and centers. For examples, we can design a loss to enforce intra-class compactness and inter-class discrepancy on the hypersphere.

Intra-Loss is designed to improve the intra-class compactness by decreasing the angle/arc between the sample and the ground truth center.
\begin{equation}
L_5=L_2 + \frac{1}{\pi} \theta_{y_i}.
\label{eq:intraloss}
\vspace{-1mm}
\end{equation}

Inter-Loss targets at enhancing inter-class discrepancy by increasing the angle/arc between different centers. 
\begin{equation}
L_6=L_2 - \frac{1}{\pi \left ( N -1\right )}  \sum_{j=1,j\neq y_i}^{N} \arccos(W^T_{y_i}W_j).
\label{eq:interloss}
\vspace{-1mm}
\end{equation}
\textcolor{black}{To enhance inter-class separability, RegularFace \cite{zhao2019regularface} explicitly distances identities by penalizing the angle between an identity and its nearest neighbor, while Minimum Hyper-spherical Energy (MHE) \cite{liu2018learning} encourages the angular diversity of neuron weights inspired by the Thomson problem. Recently, fixed classifier methods \cite{hardt2016identity,hoffer2018fix,pernici2019maximally} exhibit little or no reduction in classification performance while allowing a noticeable reduction in computational complexity, trainable parameters and communication cost. In these methods, inter-class separability is not learned but inherited from a pre-defined high-dimensional geometry \cite{pernici2019maximally}.}

Triplet-loss aims at enlarging the angle/arc margin between triplet samples. In FaceNet \cite{schroff2015facenet}, Euclidean margin is applied on the normalized features. Here, we employ the triplet-loss by the angular representation of our features as $\arccos({x_{i}^{pos} x_{i}}) + m \leq \arccos({x_i^{neg} x_{i}})$.

\subsection{Sub-center ArcFace}

Even though ArcFace has shown its power in efficient and effective face feature embedding, this method assumes that training data are clean. However, this is not true especially when the dataset is in large scale. How to enable the margin-based softmax loss to be robust to noise \textcolor{black}{is one of the main challenges impeding} the development of face recognition~\cite{wang2018devil}. In this paper, we address this problem by proposing the idea of using sub-classes for each identity, which can be directly adopted by ArcFace and will significantly increase its robustness.

\begin{figure}
\centering
\subfigure[Example of Sub-classes]{
\label{fig:visualmulticenter}
\includegraphics[height=0.15\textwidth]{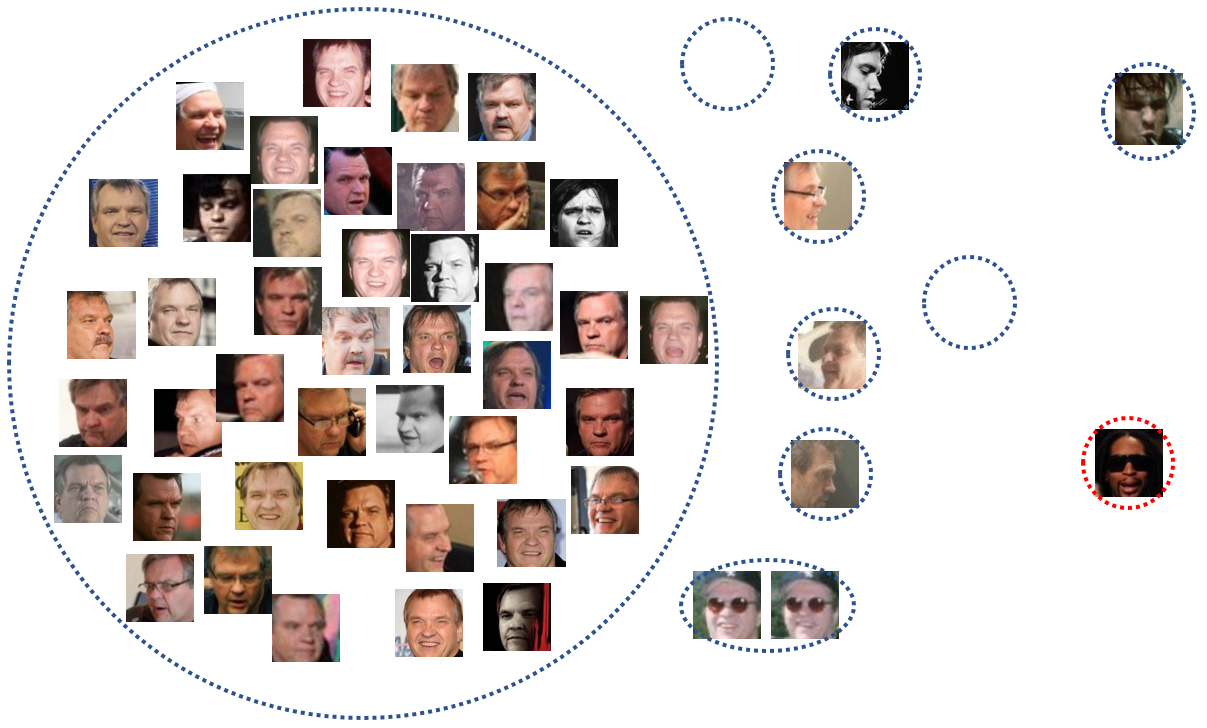}}
\subfigure[Clean Data Isolation]{
\label{fig:casiadataIsolation}
\includegraphics[height=0.15\textwidth]{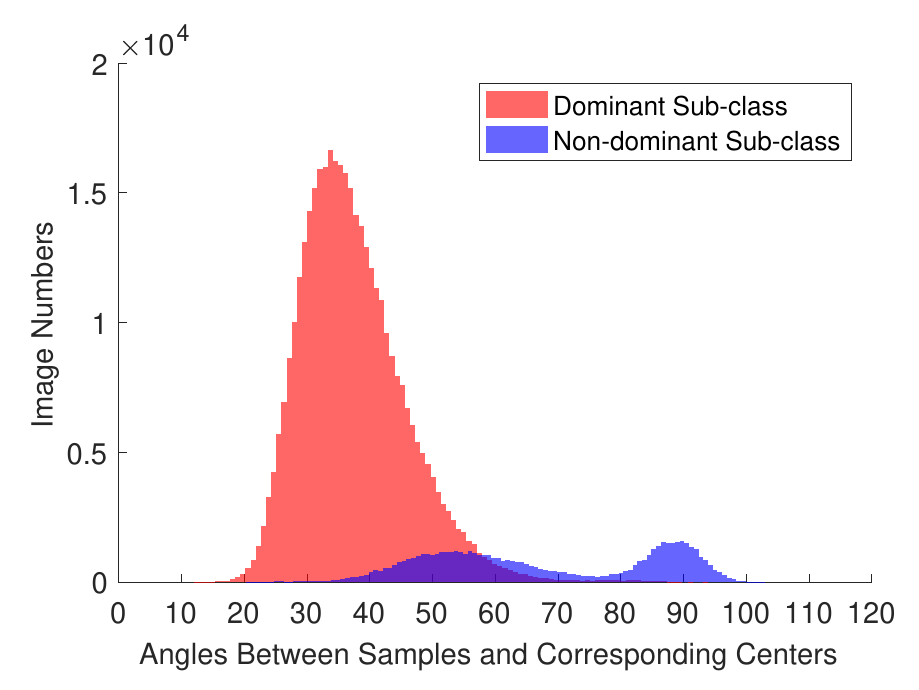}}
\caption{
(a) The sub-classes of one identity from the CASIA dataset~\cite{yi2014learning} after using the sub-center ArcFace loss ($K=10$).
Noisy samples and hard samples (\eg profile and occluded faces) are automatically separated from the majority of clean samples.
(b) Angle distribution of samples from the dominant and non-dominant sub-classes. Clean data are automatically isolated by the sub-center ArcFace.}
\vspace{-4mm}
\label{fig:multicenteroncaisa}
\end{figure}

\begin{figure}
\centering
\subfigure[{$K$=1}, All]{
\label{fig:k1}
\includegraphics[width=0.48\linewidth]{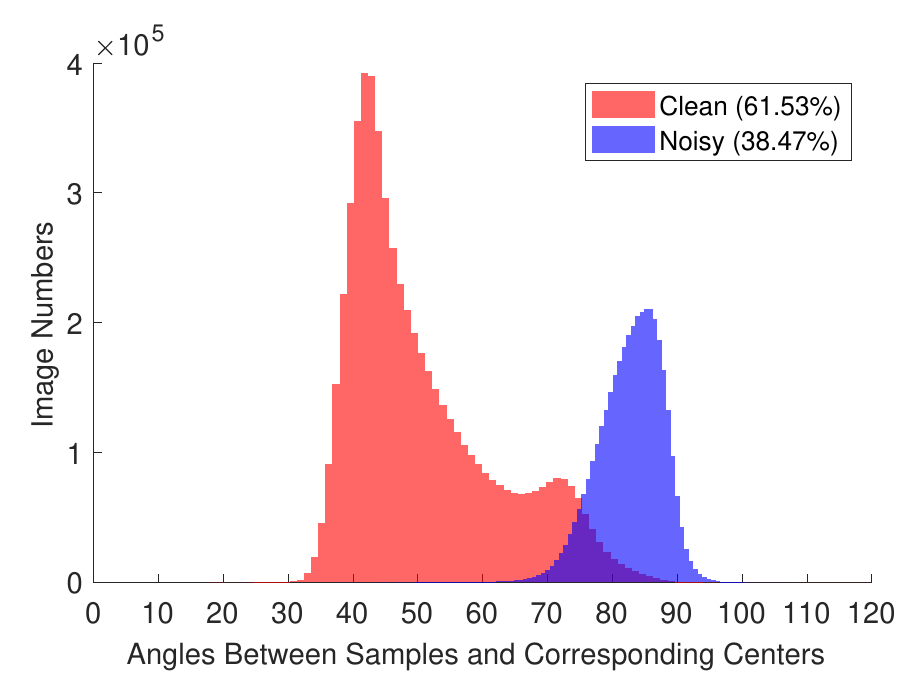}}
\subfigure[$K$=3, Dominant]{
\label{fig:k3dominantbefore}
\includegraphics[width=0.48\linewidth]{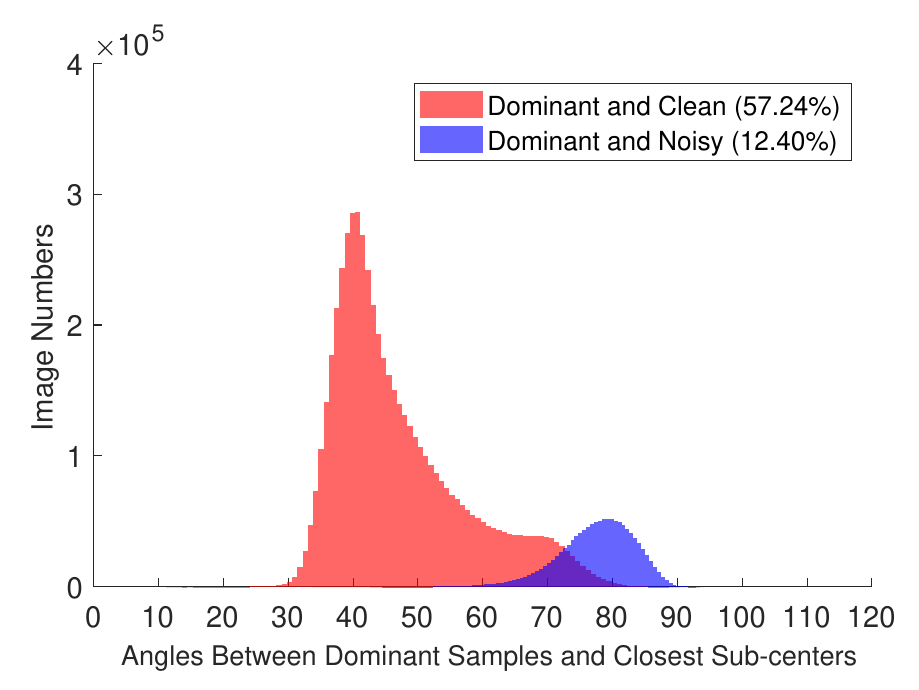}}
\subfigure[$K$=3, Non-dominant]{
\label{fig:k3nondominantbefore}
\includegraphics[width=0.48\linewidth]{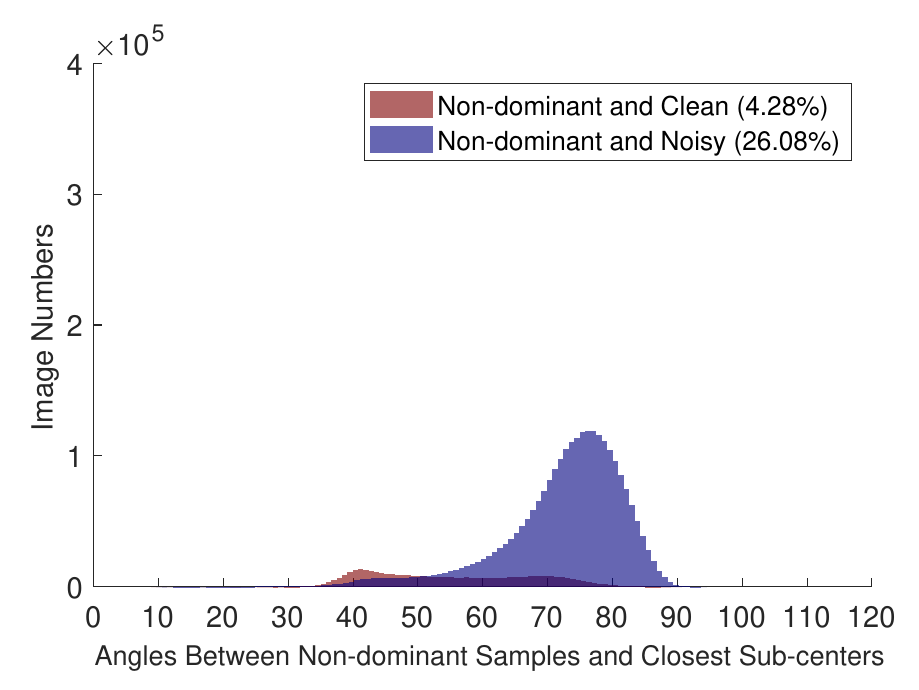}}
\subfigure[{$K=3\downarrow1$}, Non-dominant]{
\label{fig:k3nondominantafter}
\includegraphics[width=0.48\linewidth]{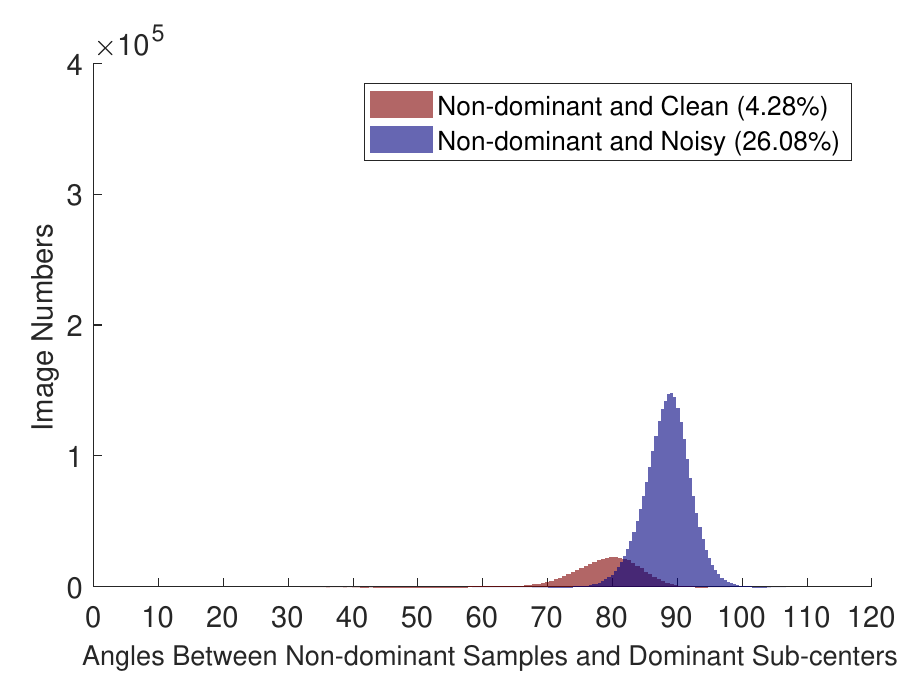}}
\caption{Data distribution of ArcFace ($K$=1) and the proposed sub-center ArcFace ($K$=3) before and after dropping non-dominant sub-centers. MS1MV0~\cite{guo2016ms} is used here. $K=3\downarrow1$ denotes sub-center ArcFace with non-dominant sub-centers dropping.}
\vspace{-4mm}
\label{fig:drop}
\end{figure}

As illustrated in Figure~\ref{fig:subcenterarcface}, we set $K$ sub-centers for each identity.  
Based on a $\ell_2$ normalization step on both embedding feature $\x_i \in \mathbb{R}^{512}$ and all sub-centers $W \in \mathbb{R}^{512 \times  N \times K }$, we get the subclass-wise similarity scores $\mathcal{S} \in \mathbb{R}^{N \times K}$ by a matrix multiplication $W^T \x_i$. Then, we employ a max pooling step on the subclass-wise similarity score $\mathcal{S} \in \mathbb{R}^{N \times K}$ to get the class-wise similarity score $\mathcal{S'} \in \mathbb{R}^{N \times 1}$.
The proposed sub-center ArcFace loss can be formulated as:
\begin{eqnarray}
{L_7}
= -\log\frac{e^{s\cos({\theta}_{y_i}+m)}}{e^{s\cos({\theta}_{y_i}+m)}+\sum_{j=1,j\neq  y_i}^{N}e^{s\cos{\theta}_{j}}},
\label{eq:subcentersarcfaceloss}
\vspace{-1mm}
\end{eqnarray}
where $\theta_{j} = arccos\left (\max_k \left ( W^T_{j_k} \x_i \right ) \right )$, $k \in \left \{ 1,\cdots, K \right \}$.

In Figure~\ref{fig:visualmulticenter}, we have visualized the clustering results of one identity from the CASIA dataset~\cite{yi2014learning} after employing the sub-center ArcFace loss ($K=10$) for training. It is obvious that the proposed sub-center ArcFace loss can automatically cluster faces such that hard samples and noisy samples are separated away from the dominant clean samples. Note that some sub-classes are empty as $K=10$ is too large for a particular identity. 
In Figure~\ref{fig:casiadataIsolation}, we show the angle distribution on the CASIA dataset~\cite{yi2014learning}. We use the pre-trained ArcFace model to predict the feature center of each identity and then calculate the angle between the sample and its corresponding feature center. As we can see from Figure~\ref{fig:casiadataIsolation}, most of the samples are close to their centers, however, there are some noisy samples which are far away from their centers. This observation on the CASIA dataset matches the noise percentage estimation ($9.3\% \sim 13.0\%$) in~\cite{wang2018devil}. To automatically obtain a clean training dataset, the noisy tail is usually removed by a hard threshold (\eg angle $\geq77^{\circ}$ or cosine $\leq 0.225$). Since sub-center ArcFace can automatically divide the training samples into dominant sub-classes and non-dominant sub-classes,  clean samples (in red) can be separated from hard and noisy samples (in blue). More specifically, the majority of clean faces ($85.6\%$) go to the dominant sub-class, while the rest hard and noisy faces go to the non-dominant sub-classes.

Even though using sub-classes can improve the robustness under noise, it undermines the intra-class compactness as hard samples are also kept away as shown in Figure~\ref{fig:casiadataIsolation}. In~\cite{guo2016ms}, MS1MV0 (around 10M images of 100K identities) is released with the estimated noise percentage around $47.1\% \sim 54.4\%$~\cite{wang2018devil}.
In~\cite{deng2019lightweight}, MS1MV0 is refined by a semi-automatic approach into a clean dataset named MS1MV3 (around 5.1M images of 93K identities). Based on these two datasets, we can get the clean and noisy labels on MS1MV0. 
In Figure~\ref{fig:k3dominantbefore} and Figure~\ref{fig:k3nondominantbefore}, we show the angle distributions of samples to their closest sub-centers (training settings: [MS1MV0, ResNet50, Sub-center ArcFace $K$=3]). In general, there are four categories of samples: (1) easy clean samples belonging to dominant sub-classes ($57.24\%$), (2) hard noisy samples belonging to dominant sub-classes ($12.40\%$), (3) hard clean samples belonging to non-dominant sub-classes ($4.28\%$), and (4) easy noisy samples belonging to non-dominant sub-classes ($26.08\%$). In Figure~\ref{fig:k1}, we show the angle distribution of samples to their corresponding centers from the ArcFace model (training settings: [MS1MV0, ResNet50, ArcFace $K$=1]). By comparing the percentages of noisy samples in Figure~\ref{fig:k3dominantbefore} and Figure~\ref{fig:k1}, we find that sub-center ArcFace can significantly decrease the noise rate to around one third (from $38.47\%$ to $12.40\%$) and this is the reason why sub-center ArcFace is more robust under noise. During the training of sub-center ArcFace, samples belonging to non-dominant sub-classes are pushed to be close to these non-dominant sub-centers as shown in Figure~\ref{fig:k3nondominantbefore}.
Since we have not set any constraint on sub-centers, the sub-centers of each identity can be quite different and even orthogonal. In Figure~\ref{fig:k3nondominantafter}, we show the angle distributions of non-dominant samples to their dominant sub-centers. By combining Figure~\ref{fig:k3dominantbefore} and Figure~\ref{fig:k3nondominantafter}, we find that the clean and noisy data have some overlaps but a constant angle threshold (between $70^{\circ}$ and $80^{\circ}$) can be easily searched to drop most of the high-confident noisy samples. 

Based on the above observations, we propose a straightforward approach to recapture intra-class compactness. We directly drop non-dominant sub-centers after the network has enough discriminative power. Meanwhile, we introduce a constant angle threshold to drop high-confident noisy data. After that, we retrain the ArcFace model from scratch on the automatically cleaned dataset.

\subsection{Inversion of ArcFace}

\begin{figure}
\centering
\includegraphics[width=0.48\textwidth]{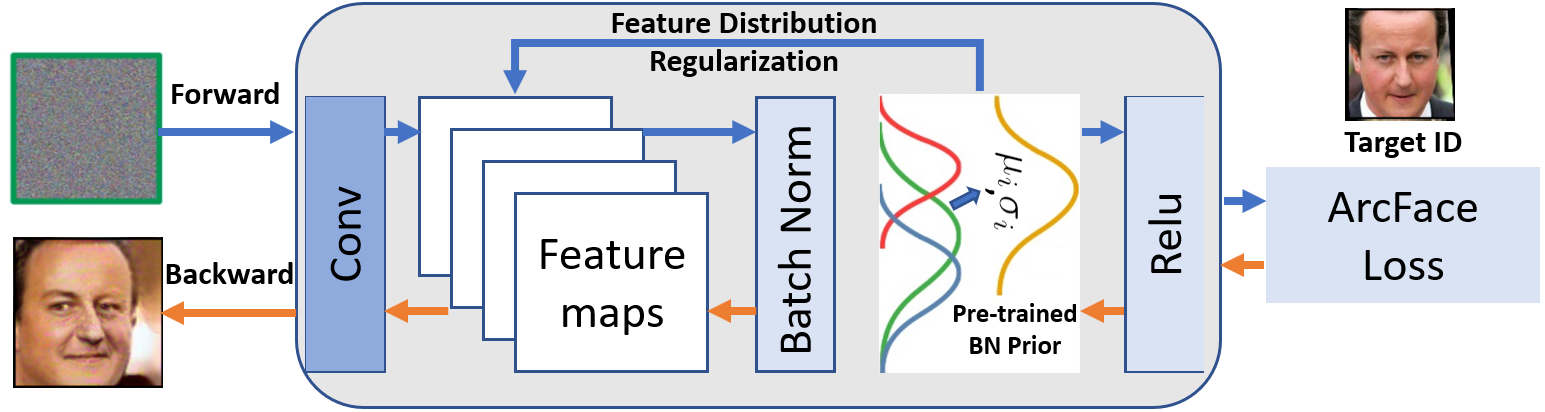}
\caption{ArcFace is not only a discriminative model but also a generative model. Given a pre-trained ArcFace model, a random input tensor can be gradually updated into a pre-defined identity by using the gradient of the ArcFace loss as well as the face statistic priors stored in the Batch Normalization layers.}
\vspace{-2mm}
\label{fig:inversearcfaceframework}
\end{figure}

\begin{algorithm}
\begin{algorithmic}
\STATE \textbf{Input:} model $\M$ with $L$ BN layers, class label $y_i$ 
\STATE \textbf{Output:} a batch of generated face images: $I^r$ 
\STATE Generate random data $I^r$ from Gaussian ($\mu=0,\sigma=1$)  
\STATE Get $\mu_i, \sigma_i$ from BN layers of $\M$, $i\in{0,\dots,L}$
\FOR{j $=1,2,\dots, T$} 
\STATE Forward propagate $\M(I^r)$ and calculate ArcFace loss
\STATE Get $\tilde\mu_i$ and $\tilde\sigma_i$ from intermediate activations, $i\in{0,\dots,L}$ 
\STATE Compute statistic loss $\min \sum_{i=0}^L \|\tilde \mu_i^r - \mu_i\|_2^2 + \|\tilde \sigma_i^r - \sigma_i\|_2^2$, 
\STATE Backward propagate and update $I^r$
\ENDFOR
\end{algorithmic}
\caption{Face Image Generation from the ArcFace Model}
\label{alg:inversearcface}
\end{algorithm}

In the above sections, we have explored how the proposed ArcFace can enhance the discriminative power of a face recognition model. In this section, we take a pre-trained ArcFace model as a white-box and reconstruct identity preserved as well as \textcolor{black}{visually plausible} face images only using the gradient of the ArcFace loss and the face statistic priors (\eg mean and variance) stored in the BN layers. 
As shown in Figure~\ref{fig:inversearcfaceframework} and illustrated in Algorithm~\ref{alg:inversearcface}, the pre-trained ArcFace model has encoded substantial information of the training distribution. The distribution, stored in BN layers via running mean and running variance, can be effectively employed to generate \textcolor{black}{visually plausible} face images, avoiding convergence outside natural faces with high confidence.

Besides the ArcFace loss (Eq.~\ref{eq:arcface}) to preserve identity, we also consider the following statistic priors during face generation:
\begin{equation}
{L_8}= \sum_{i=0}^L \|\tilde \mu_i^r - \mu_i\|_2^2 + \|\tilde \sigma_i^r - \sigma_i\|_2^2,
\label{eq:syntheticgaussian}
\vspace{-1mm}
\end{equation}
where $\mu_i^r$/$\sigma_i^r$ are the mean/standard deviation of the distribution at layer $i$, and $\mu_i$/$\sigma_i$ are the corresponding mean/standard deviation parameters stored in the $i$-th BN layer of a pre-trained ArcFace model. \textcolor{black}{
After jointly optimizing Eq.~\ref{eq:arcface} and Eq.~\ref{eq:syntheticgaussian} (${L_3} + \lambda {L_8}, \lambda = 0.05$) for $T$ steps as in Algorithm~\ref{alg:inversearcface}, we can generate faces, when fed into the network, not only have same identity as the pre-defined identity but also have a statistical distribution that closely matches the original data set.}

The above approach exploits the relationship between an input image and its class label for the reconstruction process. As the output similarity score is fixed according to predefined $N$ classes, the reconstruction is limited on images of training subjects.
To solve open-set face generation from the embedding feature, the constraints on predefined classes need to be removed. Therefore, we substitute the classification loss to the $\ell_2$ loss between feature pairs. Open-set face generation can restore the face image from any embedding feature, while close-set face generation only reconstructs face images from the class centers stored in the linear weight.

Concurrent with our work, \cite{yin2020dreaming,haroush2020knowledge,cai2020zeroq} have proposed a data-free method employing the BN priors to restore ImageNet images for distillation, quantization and pruning. Their model inversion results contain obvious artifact in the background due to the translation augmentation during training. By contrast, our ArcFace model is trained on normalized face crops without background, thus the restored faces exhibit less artifact. Besides, these data-free methods only considered close-set image generation but ArcFace can freely restore both close-set and open-set subjects. In this paper, we show that the proposed additive angular margin loss can also improve face generation. 

\section{Experiments}

\subsection{Implementation Details}

\begin{table}[t!]
\begin{center}
\caption{Face datasets for training and testing. ``(D)'' refers to the distractors. IBUG-500K is the training data automatically refined by the proposed sub-center ArcFace. LFR2019-Image and LFR2019-Video are the proposed large-scale image and video test sets.}
\vspace{-4mm}
\label{table:dataset}
\small
\begin{tabular}{c|c|c}
\hline
 Datasets   & \#Identity & \#Image/Video\\
\hline
CASIA \cite{yi2014learning} & 10K  & 0.5M  \\
VGG2 \cite{cao2017vggface2}     & 9.1K & 3.3M  \\
MS1MV0      \cite{guo2016ms}       & 100K & 10M \\
MS1MV3     \cite{deng2019lightweight} & 93K  & 5.1M \\ 
Celeb500K  \cite{cao2018celeb}     & 500K & 50M \\ 
{\bf IBUG-500K}                    & 493K & 11.96M \\
\hline\hline
LFW \cite{huang2007labeled}   & 5,749  & 13,233 \\
YTF  \cite{wolf2011face}          & 1,595  & 3,425 \\ 
CFP-FP \cite{sengupta2016frontal} & 500 & 7,000\\
CPLFW \cite{zheng2018cross} & 5,749  & 11,652 \\
AgeDB \cite{Moschoglou2017AgeDB} & 568 & 16,488 \\
CALFW \cite{zheng2017cross} & 5,749  & 12,174 \\
MegaFace \cite{kemelmacher2016megaface} & 530 & 1M (D) \\
IJB-B    \cite{whitelam2017iarpa}       & 1,845   & 76.8K \\
IJB-C    \cite{maze2018iarpa}           & 3,531   & 148.8K \\
{\bf LFR2019-Image} \cite{deng2019lightweight}  & 5.7K   & 1.58M(D)\\
{\bf LFR2019-Video} \cite{deng2019lightweight} & 10K    & 200K\\
\hline
\end{tabular}
\end{center}
\vspace{-4mm}
\end{table}

\begin{figure*}
\centering
\subfigure[Gender]{
\label{fig:gender}
\includegraphics[width=0.18\textwidth]{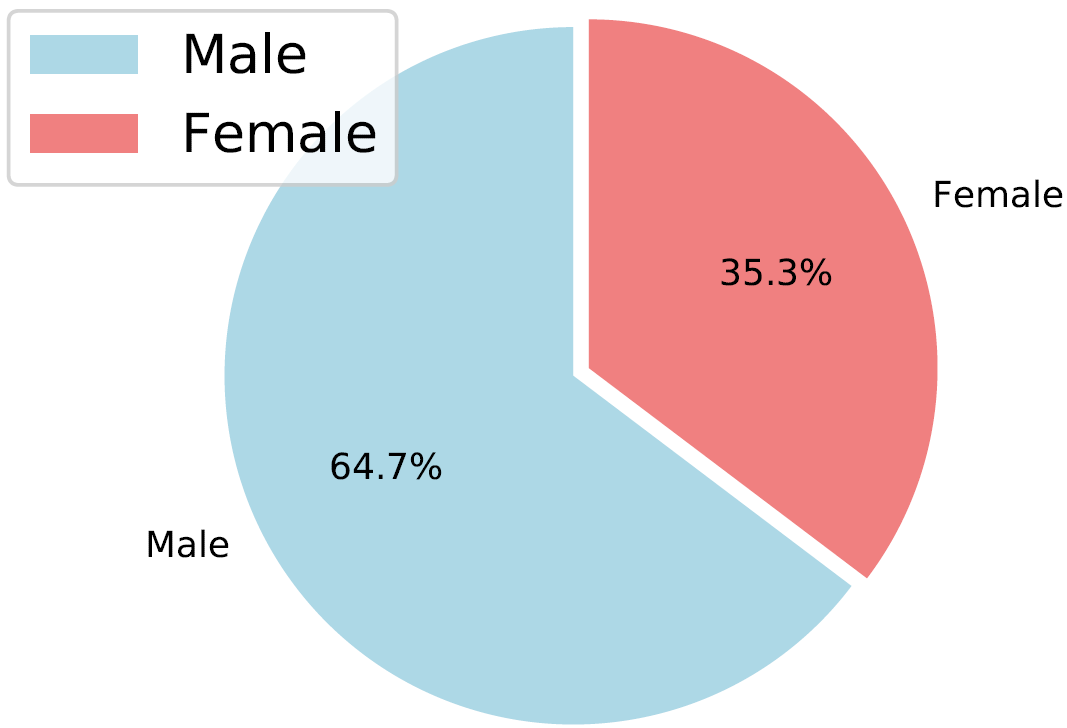}}
\subfigure[Race]{
\label{fig:race}
\includegraphics[width=0.18\textwidth]{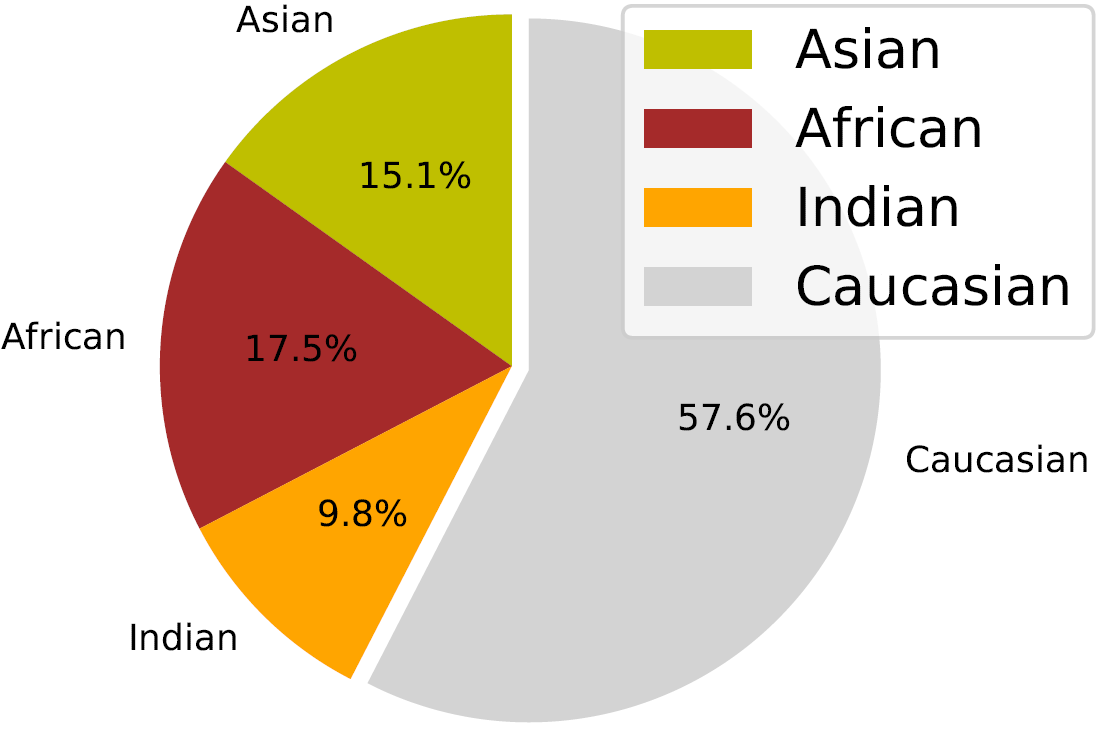}}
\subfigure[Pose (Yaw)]{
\label{fig:pose}
\includegraphics[width=0.18\textwidth]{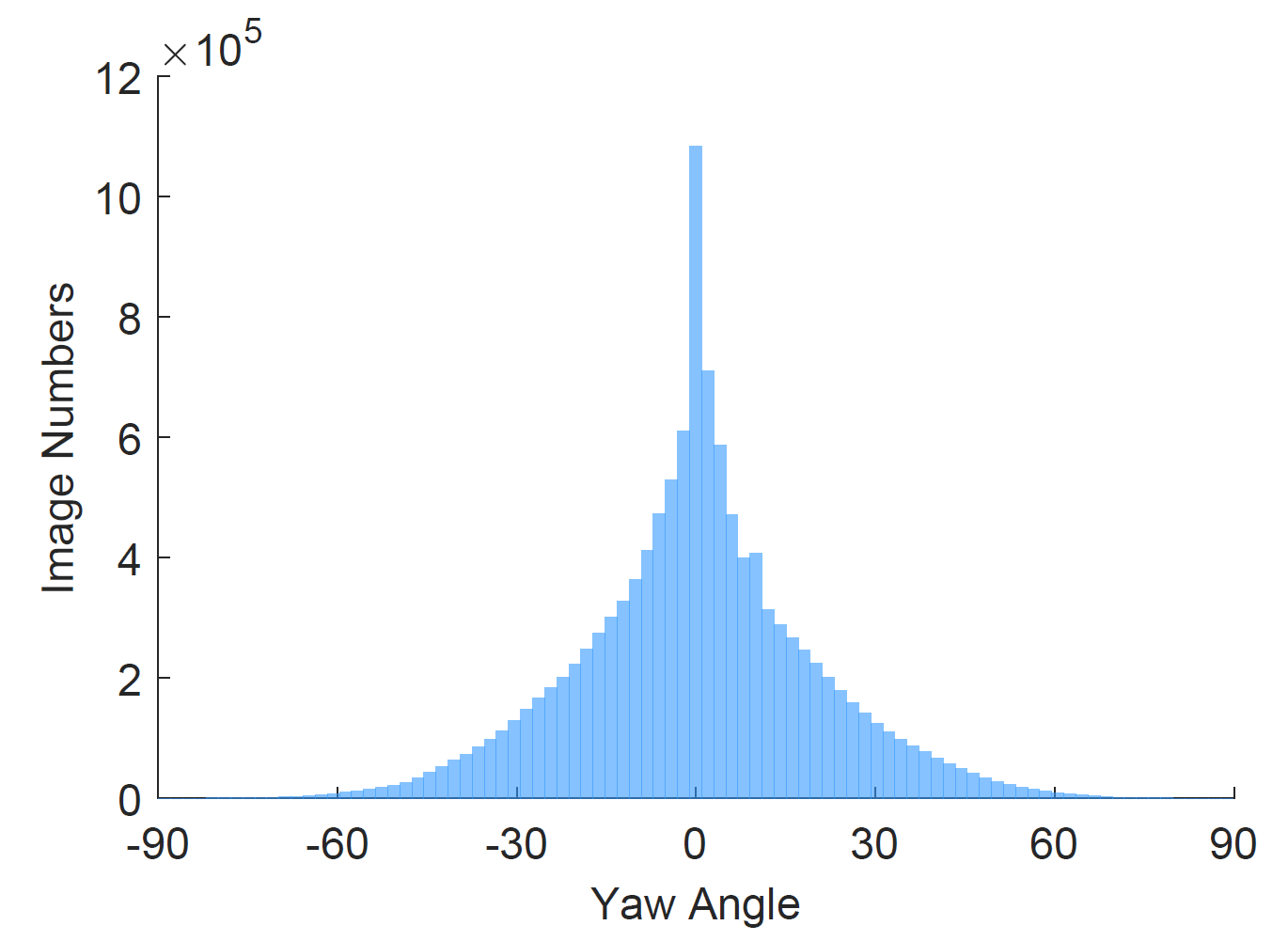}}
\subfigure[Age]{
\label{fig:age}
\includegraphics[width=0.18\textwidth]{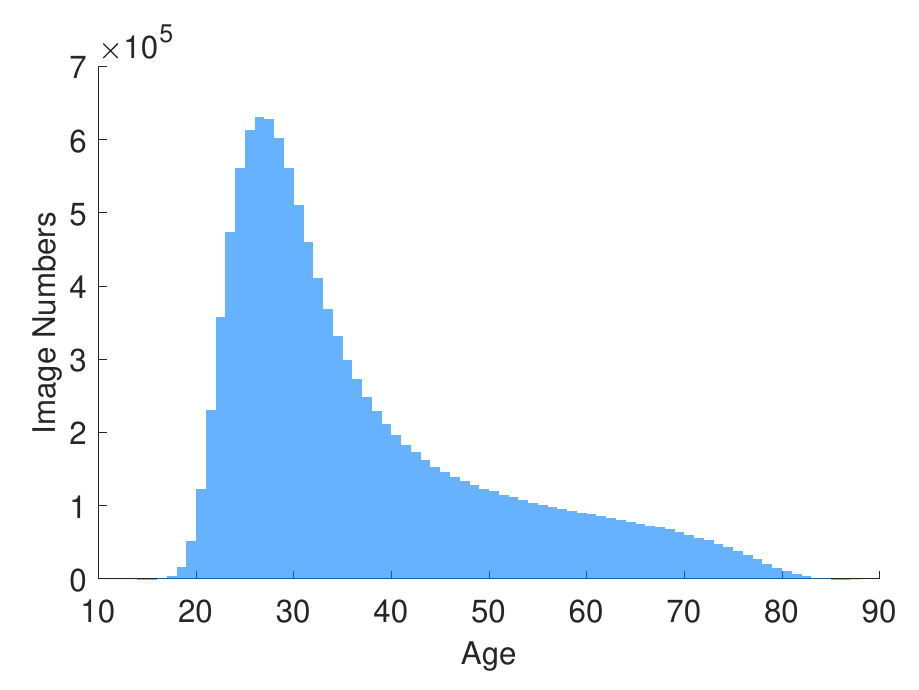}}
\subfigure[Image Number]{
\label{fig:longtail}
\includegraphics[width=0.18\textwidth]{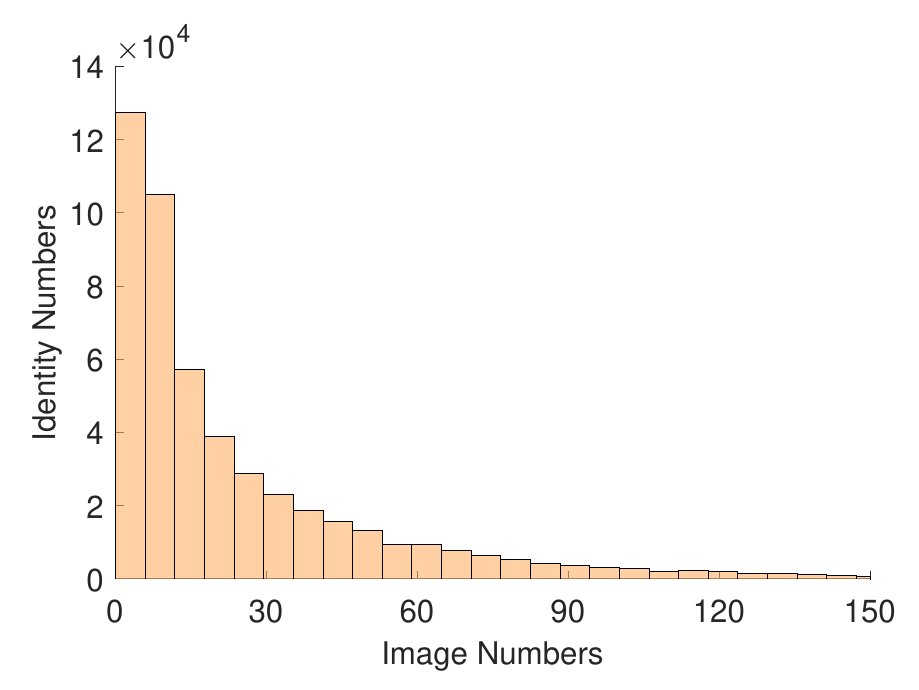}}
\caption{IBUG-500K statistics. We show the (a) gender, (b) race, (c) yaw pose, (d) age and (e) image number distributions of the proposed large-scale training dataset.}
\vspace{-4mm}
\label{fig:ibug500kstatistics}
\end{figure*}

\noindent {\bf Training Datasets.} As given in Table \ref{table:dataset}, we separately employ CASIA \cite{yi2014learning}, VGG2 \cite{cao2017vggface2}, MS1MV0 \cite{guo2016ms} and Celeb500K \cite{cao2018celeb} as our training data in order to conduct fair comparison with other methods. 
MS1MV0 (loose cropped version) \cite{guo2016ms} is a raw data with the estimated noise percentage around $47.1\% \sim 54.4\%$ ~\cite{wang2018devil}. MS1MV3~\cite{deng2019lightweight} is cleaned from MS1MV0 \cite{guo2016ms} by a semi-automatic approach. We employ ethnicity-specific annotators \textcolor{black}{(\eg African, Caucasian, Indian and Asian)} for large-scale face image annotations, as the boundary cases (\eg hard samples and noisy samples) are very hard to distinguish if the annotator is not familiar with the identity. Celeb500K~\cite{cao2018celeb} is collected in the same way as MS1MV0~\cite{guo2016ms}, using the celebrity name list~\cite{guo2016ms} to search identities from Google and download the top-ranked face images. We download 25M images of 500K identities, and employ RetinaFace \cite{deng2020retinaface} to detect faces larger than $50\times50$ from the original images. By employing the proposed sub-center ArcFace, we can automatically clean MS1MV0 \cite{guo2016ms} and Celeb500K~\cite{cao2018celeb}. After removing the overlap identities (about 50K) through the ID string, we combine the automatically cleaned MS1MV0 and Celeb500K and obtain a large-scale face image dataset, named IBUG-500K, including 11.96 million images of 493K identities. Figure \ref{fig:ibug500kstatistics}
illustrates the gender, race, pose, age and image number distributions of the proposed IBUG-500K dataset.

\noindent {\bf Test Datasets.} During training, we explore efficient face verification datasets (\eg LFW \cite{huang2007labeled}, CFP-FP \cite{sengupta2016frontal}, AgeDB \cite{Moschoglou2017AgeDB}) to check the convergence status of the model. Besides the most widely used LFW \cite{huang2007labeled} and YTF \cite{wolf2011face} datasets, we also report the performance of ArcFace on the recent datasets (\eg CPLFW \cite{zheng2018cross} and CALFW \cite{zheng2017cross}) with large pose and age variations. We also extensively test the proposed ArcFace on large-scale image datasets (\eg MegaFace \cite{kemelmacher2016megaface}, IJB-B \cite{whitelam2017iarpa}, IJB-C \cite{maze2018iarpa} and LFR2019-Image \cite{deng2019lightweight}) and large-scale video datasets (LFR2019-Video \cite{deng2019lightweight}). Detailed dataset statistics are presented in Table \ref{table:dataset}. For the LFR2019-Image dataset, there are 274K images from the 5.7K LFW identities \cite{huang2007labeled} and 1.58M distractors downloaded from Flickr. For the LFR2019-Video dataset, there are 200K videos of 10K identities collected from various shows, films and television dramas. The length of each video ranges from $1$ to $30$ seconds. Both the LFR2019-Image dataset and the LFR2019-Video dataset are manually cleaned to ensure the unbiased evaluation of different face recognition models. 

\noindent {\bf Experimental Settings.} For data prepossessing, we follow the recent papers \cite{liu2017sphereface,tencent2017CosineFace} to generate the normalized face crops ($112\times112$) by utilizing five facial points predicted by RetinaFace \cite{deng2020retinaface}. For the embedding network, we employ the widely used CNN architectures, ResNet50 and ResNet100 \cite{he2016deep,han2016deep} without the bottleneck structure. After the last convolutional layer, we explore the BN \cite{ioffe2015batch}-Dropout \cite{srivastava2014dropout}-FC-BN structure to get the final $512$-$D$ embedding feature. In this paper, we use ([training dataset, network structure, loss]) to facilitate understanding of different experimental settings. 

We follow \cite{tencent2017CosineFace} to set the feature scale $s$ to 64 and choose the angular margin $m$ of ArcFace at $0.5$. All recognition experiments in this paper are implemented by MXNet \cite{chen2015mxnet}. \textcolor{black}{We set the batch size to $512$ and train models on eight NVIDIA Tesla P40 (24GB) GPUs. We set the momentum to $0.9$ and weight decay to $5e-4$. For the ArcFace training, we employ the SGD optimizer and follow \cite{tencent2017CosineFace,cao2017vggface2} to design the learning rate schedules for different datasets. On CASIA, the learning rate starts from $0.1$ and is divided by $10$ at 20, 28 epochs. The training process is finished at 32 epochs. On VGG2, the learning rate is decreased at 6, 9 epochs and we finish training at 12 epochs. On MS1MV3 and IBUG-500K, we refer to the verification accuracy on CFP-FP and AgeDB to reduce the learning rate at 8, 14 epochs and terminate at 18 epochs.}

For the training of the proposed sub-center ArcFace on MS1MV0 \cite{guo2016ms}, we also employ the same learning rate schedule as on MS1MV3 to train the first round of model ($K$=3). Then, we drop non-dominant sub-centers ($K=3\downarrow1$) and high-confident noisy data ($>75^{\circ}$) by using the first round model through an off-line way. Finally, we retrain the model from scratch using the automatically cleaned data. For the experiments of the sub-center ArcFace on Celeb500K \cite{cao2018celeb}, the only difference is the learning rate schedule, which is same as on IBUG-500K. 

During testing of the face recognition models, we only keep the feature embedding network without the fully connected layer (160MB for ResNet50 and 250MB for ResNet100) and extract the $512$-$D$ features ($8.9$ ms/face for ResNet50 and $15.4$ ms/face for ResNet100) for each normalized face. To get the embedding features for templates (\eg IJB-B and IJB-C) or videos (\eg YTF and LFR2019-Video), we simply calculate the feature center of all images from the template or all frames from the video. 

\subsection{Ablation Study on ArcFace}

\begin{table}[t!]
\begin{center}
\caption{Verification results ($\%$) of different loss functions ([CASIA, ResNet50, Loss*]).}
\vspace{-4mm}
\label{table:losscompare}
\small
\begin{tabular}{c|c|c|c}
\hline
Loss Functions   & LFW & CFP-FP & AgeDB \\
\hline
ArcFace (0.4)         & 99.53 & 95.41 & 94.98   \\
ArcFace (0.45)        & 99.46 & 95.47 & 94.93 \\
ArcFace (0.5)         & {\bf 99.53} & {\bf 95.56} & {\bf 95.15}\\
ArcFace (0.55)        & 99.41  & 95.32   & 95.05   \\
\hline
SphereFace \cite{liu2017sphereface}  & 99.42 & - & - \\
SphereFace (1.35)                    & 99.11 & 94.38 & 91.70 \\
CosFace \cite{tencent2017CosineFace} & 99.33 & - & - \\
CosFace (0.35)                       & 99.51 & 95.44 & 94.56 \\
\hline
CM1 (1, 0.3, 0.2)                    & 99.48 & 95.12 & 94.38 \\
CM2 (0.9, 0.4, 0.15)                 & 99.50 & 95.24 & 94.86 \\
\hline
Softmax              & 99.08 & 94.39 & 92.33 \\
Norm-Softmax ($s=64$) & 98.56 & 89.79 & 88.72 \\
Norm-Softmax ($s=20$)  & 99.20   & 94.61   & 92.65 \\
Norm-Softmax+Intra          &  99.30  & 94.85  & 93.58   \\
Norm-Softmax+Inter          &  99.22  & 94.73  & 92.94   \\
Norm-Softmax+Intra+Inter        &  99.31  & 94.88  & 93.76 \\
Triplet (0.35)       & 98.98 & 91.90 & 89.98 \\
ArcFace+Intra        & 99.45 & 95.37 & 94.73 \\
ArcFace+Inter        & 99.43 & 95.25 & 94.55 \\
ArcFace+Intra+Inter  & 99.43 & 95.42 & 95.10 \\
ArcFace+Triplet      & 99.50 & 95.51 & 94.40  \\
\hline
\end{tabular}
\end{center}
\vspace{-4mm}
\end{table}

In Table \ref{table:losscompare}, we first explore the angular margin setting for ArcFace on the CASIA dataset with ResNet50. The best margin observed in our experiments is $0.5$. Using the proposed combined margin framework in Eq. \ref{eq:combineloss}, it is easier to set the margin of SphereFace and CosFace which we find to have optimal performance when setting at $1.35$ and $0.35$, respectively.  
Our implementations for both SphereFace and CosFace can lead to excellent performance without observing any difficulty in convergence. The proposed ArcFace achieves the highest verification accuracy on all three test sets. In addition, we perform extensive experiments with the combined margin framework (some of the best performance is observed for CM1 (1, 0.3, 0.2) and CM2 (0.9, 0.4, 0.15)) guided by the target logit curves in Figure \ref{fig:soatargetlogit}. The combined margin framework leads to better performance than individual SphereFace and CosFace but upper-bounded by the performance of ArcFace. 

\textcolor{black}{
Besides the comparison with margin-based methods, we conduct a further comparison between ArcFace and other losses which aim at enforcing intra-class compactness (Eq. \ref{eq:intraloss}) and inter-class discrepancy (Eq. \ref{eq:interloss}). As the baseline, we choose the softmax loss. After weight and feature normalization, we have observed obvious performance drops on CFP-FP and AgeDB with the feature re-scale parameter $s$ set as $64$. To obtain comparable performance as the softmax loss, we have searched the best scale parameter $s=20$ for Norm-Softmax.
By combining the Norm-Softmax with the intra-class loss, the performance improves on CFP-FP and AgeDB. However, combining the Norm-Softmax with the inter-class loss only slightly improves the accuracy. Employing margin penalty within triplet samples is less effective than inserting margin between samples and centers as in ArcFace, indicating local comparisons in the Triplet loss are not as effective as global comparisons in ArcFace. Finally, we incorporate the Intra-loss, Inter-loss and Triplet-loss into ArcFace, but no obvious improvement is observed, which leads us to believe that ArcFace is already enforcing intra-class compactness, inter-class discrepancy and classification margin.}

\subsection{Ablation Study on Sub-center ArcFace}

\begin{table}[t!]
\begin{center}
\caption{Ablation experiments of different settings of the proposed sub-center ArcFace on MS1MV0, MS1MV3 and Celeb500K. The $1$:$1$ verification accuracy (TPR@FPR=$1e$$-$$4$) is reported on the IJB-B and IJB-C datasets. ([MS1MV0 / MS1MV3 / Celeb500K, ResNet50, Sub-center ArcFace])}
\vspace{-4mm}
\label{tab:ms1mv0andv3}
\small
\begin{tabular}{l|cc}
\hline
Settings     &  IJB-B & IJB-C\\ 
\hline
(1) MS1MV0,$K$=1 & 87.87  & 90.27 \\ 
\hline
(2) MS1MV0,$K$=3 & 91.70  & 93.72  \\
(3) MS1MV0,$K$=3, softmax pooling~\cite{qian2019softtriple}  & 91.53  & 93.55  \\
\hline
(4) MS1MV0,$K$=5  & 91.47  & 93.62  \\ 
(5) MS1MV0,$K$=10 & 63.84  & 67.94 \\
\hline
(6) MS1MV0, $K=3\downarrow1$, drop $>70^{\circ}$    & 94.44  & 95.91  \\ 
(7) MS1MV0, $K=3\downarrow1$, drop $>75^{\circ}$    & 94.56  & 95.92  \\ 
(8) MS1MV0, $K=3\downarrow1$, drop $>80^{\circ}$    & 94.04  & 95.74  \\ 
(9) MS1MV0, $K=3\downarrow1$, drop $>85^{\circ}$    & 93.33  & 95.10  \\
(10) MS1MV0, $K$=3, regularizer~\cite{qian2019softtriple}  & 91.53  & 93.64 \\ 
\hline
(11) MS1MV0,Co-mining~\cite{wang2019co} & 91.80 & 93.82  \\  
(12) MS1MV0,NT~\cite{hu2019noise}       & 91.57 & 93.65  \\ 
(13) MS1MV0,NR~\cite{Zhong2019Unequal}  & 91.58 & 93.60  \\ 
\hline
(14) MS1MV3, $K$=1      & 95.13  & 96.50  \\
(15) MS1MV3, $K$=3      & 94.84  & 96.35  \\ 
(16) MS1MV3, $K=3\downarrow1$  & 94.87  & 96.43 \\ 
\hline
(17) Celeb500K, $K$=1      & 90.96 & 92.15 \\ 
(18) Celeb500K, $K$=3      & 93.76  & 94.90 \\ 
(19) Celeb500K, $K=3\downarrow1$  & {\bf 95.65}  & {\bf 96.91} \\ 
\hline
\end{tabular}
\end{center}
\vspace{-4mm}
\end{table}

In Table~\ref{tab:ms1mv0andv3}, we conduct extensive experiments to investigate the proposed sub-center ArcFace on noisy training data (\eg MS1MV0 \cite{guo2016ms} and Celeb500K \cite{cao2018celeb}). Models trained on the manually cleaned MS1MV3~\cite{deng2019lightweight} are taken as the reference. We train ResNet50 networks under different settings and evaluate the performance by adopting TPR@FPR=1e-4 on IJB-C, which is more objective and less affected by the noise within the test data~\cite{xie2018comparator}. 

From Table~\ref{tab:ms1mv0andv3}, we have the following observations:
\begin{itemize}
\item  ArcFace has an obvious performance drop (from (14) $96.50\%$ to (1) $90.27\%$) when the training data is changed from the clean MS1MV3 to the noisy MS1MV0. By contrast, sub-center ArcFace is more robust ((2) $93.72\%$) under massive noise. 
\item  \textcolor{black}{Too many sub-centers (too large $K$) can obviously undermine the intra-class compactness and decrease the accuracy (from (2) $93.72\%$ to (5) $67.94\%$). This observation indicates that noise tolerance and intra-class compactness should be balanced during training. Considering the GPU memory consumption, we select $K$=3 in this paper.}
\item  The nearest sub-center assignment by the max pooling is slightly better than the softmax pooling~\cite{qian2019softtriple} ((2) $93.72\%$ vs. (3) $93.55\%$). Thus, we choose the more efficient max pooling operator in the following experiments.
\item  Dropping non-dominant sub-centers and high-confident noisy samples can achieve better performance than adding regularization~\cite{qian2019softtriple} to enforce compactness between sub-centers ((7) $95.92\%$ vs. (10) $93.64\%$). Besides, \textcolor{black}{the} performance of our method is not very sensitive to the constant threshold ((6) $95.91\%$, (7) $95.92\%$ and (8) $95.74\%$), and we select $75^{\circ}$ as the threshold for dropping high-confident noisy samples in the following experiments. 
\item  Co-mining~\cite{wang2019co} and re-weighting methods~\cite{hu2019noise,Zhong2019Unequal} can also improve the robustness under massive noise, but sub-center ArcFace can do better through automatic clean and noisy data isolation during training ((7) $95.92\%$ vs. (11) $93.82\%$, (12) $93.65\%$ and (13) $93.60\%$). 
\item  On the clean dataset (MS1MV3), sub-center ArcFace achieves similar performance as ArcFace ((16) $96.43\%$ vs. (14) $96.50\%$). By employing the threshold of $75^{\circ}$ on MS1MV3, $4.18\%$ hard samples are removed, but the performance only slightly decreases, thus we estimate MS1MV3 still contains some noises.
\item  The proposed sub-center ArcFace trained on noisy MS1MV0 can achieve comparable performance compared to ArcFace trained on manually cleaned MS1MV3 ((7) $95.92\%$ vs. (14) $96.50\%$). 
\item By enlarging the training data, sub-center ArcFace can easily achieve better performance even though it is trained from noisy web faces ((19) $96.91\%$ vs. (13) $96.50\%$).
\end{itemize}

\subsection{Benchmark Results}

\begin{table}[t!]
\begin{center}
\caption{Verification performance ($\%$) of different methods on LFW and YTF. ([Dataset*, ResNet100, ArcFace])} 
\vspace{-4mm}
\label{table:lfwytf}
\small
\begin{tabular}{c|c|c|c}
\hline
Method &  \#Image &  LFW     & YTF\\
\hline
DeepID \cite{sun2014deep}            & 0.2M    &99.47 &93.20\\
Deep Face \cite{taigman2014deepface} & 4.4M    &97.35 &91.4\\
VGG Face \cite{parkhi2015deep}       & 2.6M    &98.95 &97.30\\
FaceNet \cite{schroff2015facenet}    & 200M    &99.63 &95.10\\
Baidu \cite{liu2015targeting}     & 1.3M       &99.13 & -\\
Center Loss \cite{wen2016discriminative} & 0.7M &99.28 &94.9\\
Range Loss \cite{zhang2016range}     &5M       &99.52 &93.70\\
Marginal Loss \cite{deng2017marginal} & 3.8M   &99.48 &95.98 \\
SphereFace \cite{liu2017sphereface}  & 0.5M     &99.42 &95.0\\
SphereFace+ \cite{liu2018learning}   & 0.5M    &99.47 & - \\
CosFace \cite{tencent2017CosineFace} & 5M    &99.73 &97.6\\
RegularFace \cite{zhao2019regularface}  &3.1M  &  99.61 & 96.7 \\ 
UniformFace \cite{duan2019uniformface}   & 6.1M  & 99.8 & 97.7\\
DAL \cite{wang2019decorrelated}         & 0.5M      & 99.47 &- \\
FTL \cite{yin2019feature}                & 5M & 99.55 & -\\
Fair Loss \cite{liu2019fair} & 0.5M & 99.57 & 96.2\\
Unequal-training \cite{Zhong2019Unequal} & 0.55M & 99.53 & 96.04 \\
Noise-Tolerant \cite{hu2019noise} & 1M noisy & 99.72 & 97.36 \\
AdaptiveFace \cite{liu2019adaptiveface} & 5M & 99.62& -\\
AFRN \cite{kang2019attentional} & 3.1M & {\bf 99.85} & 97.1\\
PFE \cite{shi2019probabilistic} &4.4M &  99.82 & 97.36\\
DUL \cite{chang2020data} &3.6M &99.78 &96.78 \\
RDCFace \cite{zhao2020rdcface} & 1.7M & 99.80 & 97.10\\
HPDA \cite{wang2020hierarchical} & 5M & 99.80 &-\\
URFace \cite{shi2020towards} & 5M & 99.78 & -\\
CircleLoss \cite{sun2020circle} & 3.6M & 99.73 & 96.38 \\
GroupFace \cite{kim2020groupface} &5.8M & {\bf 99.85} & 97.8 \\
BioMetricNet \cite{ali2020biometricnet} & 3.8M & 99.80  & {\bf 98.06} \\
BroadFace \cite{kim2020broadface} &5.8M &{\bf 99.85} & 98.0 \\
\textcolor{black}{IBUG500K,R100,BroadFace} & 11.96M   & 99.83 & 98.03 \\
\hline
\textcolor{black}{MS1MV3, R100, ArcFace}              & 5.1M   & 99.83 & 98.02 \\
IBUG500K, R100, ArcFace             & 11.96M   & 99.83 & 98.01 \\
\hline
\end{tabular}
\end{center}
\vspace{-4mm}
\end{table}

\begin{table}[t!]
\begin{center}
\caption{Verification performance ($\%$) of different methods on CFP-FP, CPLFW, AgeDB and CALFW. ([Dataset*, ResNet100, ArcFace])}
\vspace{-4mm}
\label{table:calfwandcplfw}
\small
\resizebox{\linewidth}{!}{
\begin{tabular}{c|cc|cc}
\hline
Method                                  & CFP-FP & CPLFW & AgeDB & CALFW \\
\hline
Center Loss \cite{wen2016discriminative}&- & 77.48 & - & 85.48   \\
SphereFace \cite{liu2017sphereface}     &- & 81.40 & - & 90.30  \\
VGGFace2  \cite{cao2017vggface2}        &- & 84.00 & - & 90.57    \\
MV-Softmax \cite{wang2019mis}           & 98.28 & 92.83 & 97.95 & {\bf 96.10} \\
Search-Softmax \cite{wang2020loss}    & 95.64 & 89.50  & 97.75 & 95.40 \\
FaceGraph \cite{zhang2020global}     & 96.90 & 92.27 &97.92 & 95.67 \\ 
CurricularFace \cite{huang2020curricularface} & 98.36 & 93.13 & 98.37 & 96.05 \\
\hline
\textcolor{black}{MS1MV3, R100, ArcFace}               & 98.79 & 93.21 & 98.23 & 96.02\\
IBUG500K, R100, ArcFace                              & {\bf 98.87} & {\bf 93.43} &{\bf 98.38} & {\bf 96.10}\\
\hline
\end{tabular}}
\end{center}
\vspace{-4mm}
\end{table}

\begin{figure}
\centering
\subfigure[LFW ($99.83\%$)]{
\label{fig:lfwhist}
\includegraphics[width=0.23\textwidth]{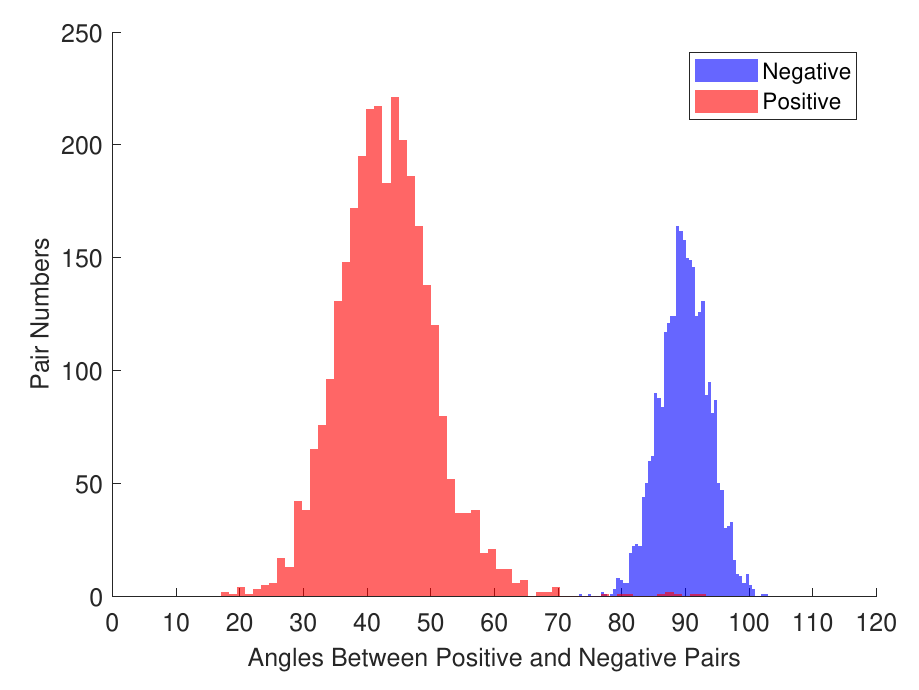}}
\subfigure[YTF ($98.01\%$)]{
\label{fig:YTFhist}
\includegraphics[width=0.23\textwidth]{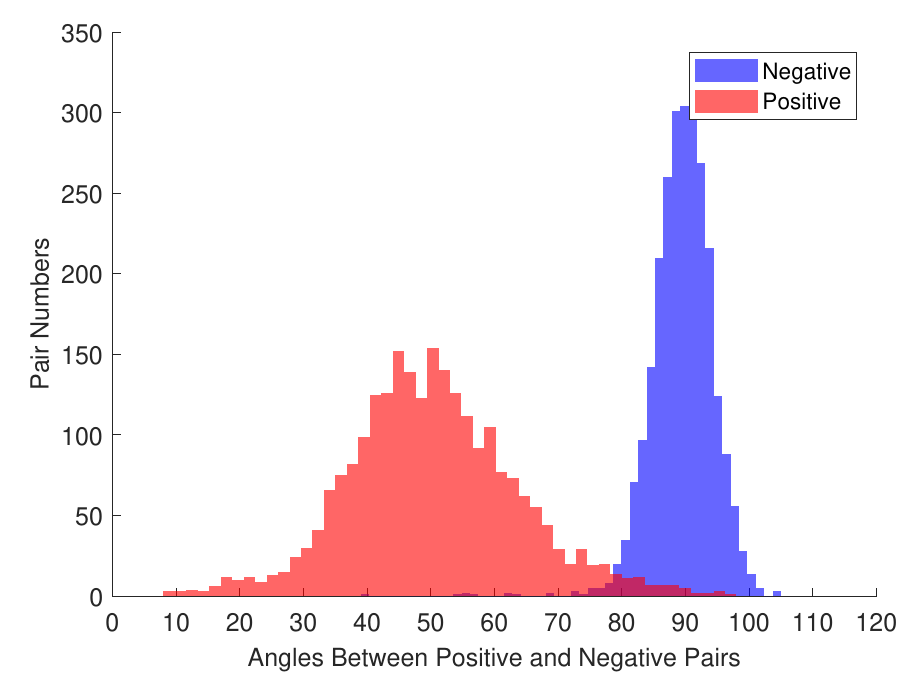}}
\subfigure[CFP-FP ($98.87\%$)]{
\label{fig:cfpfphist}
\includegraphics[width=0.23\textwidth]{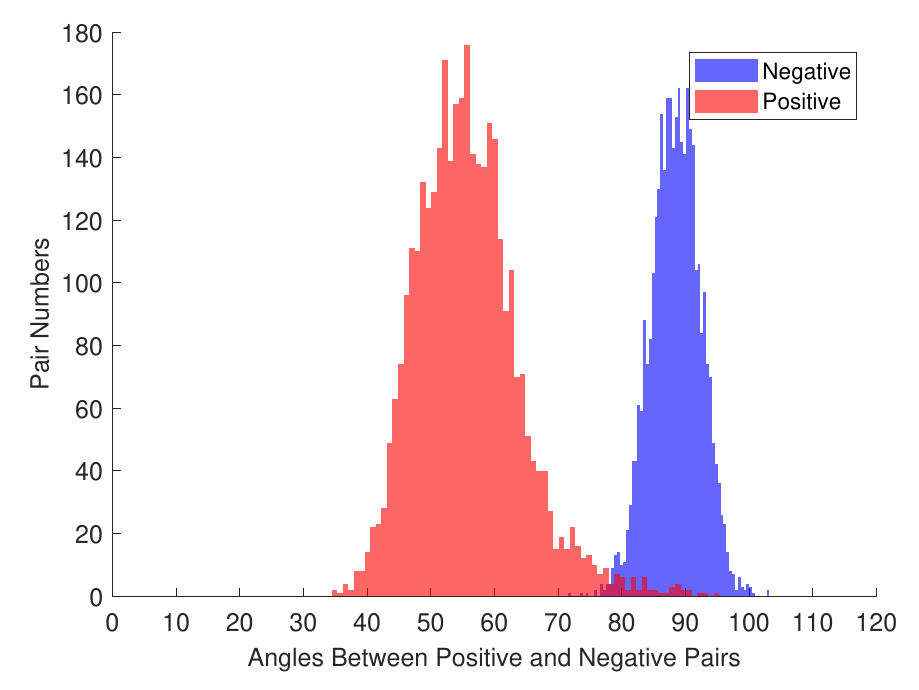}}
\subfigure[CPLFW ($93.43\%$)]{
\label{fig:CPLFWhist}
\includegraphics[width=0.23\textwidth]{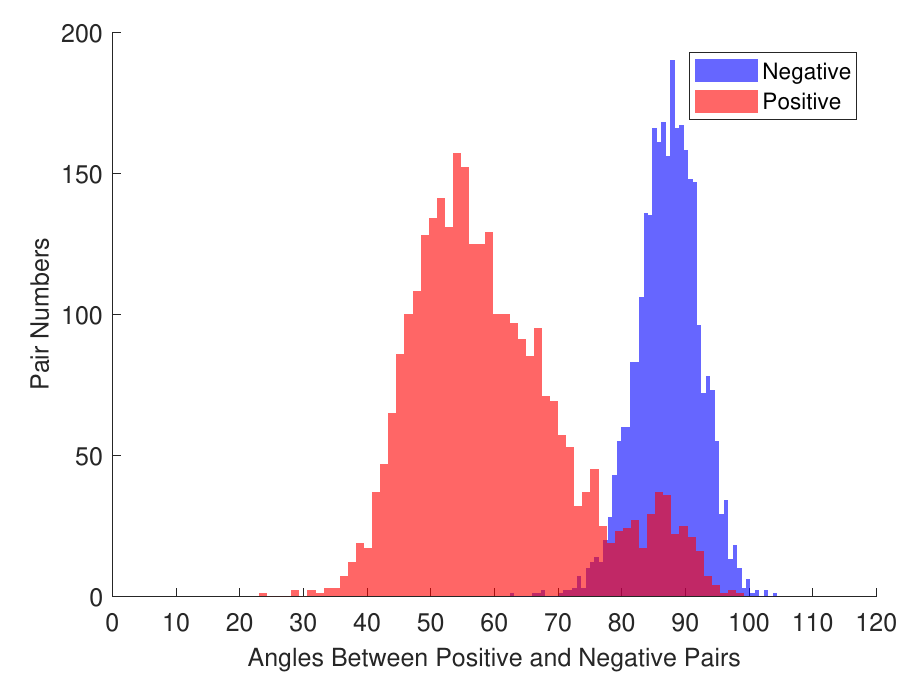}}
\subfigure[AgeDB ($98.38\%$)]{
\label{fig:AgeDBhist}
\includegraphics[width=0.23\textwidth]{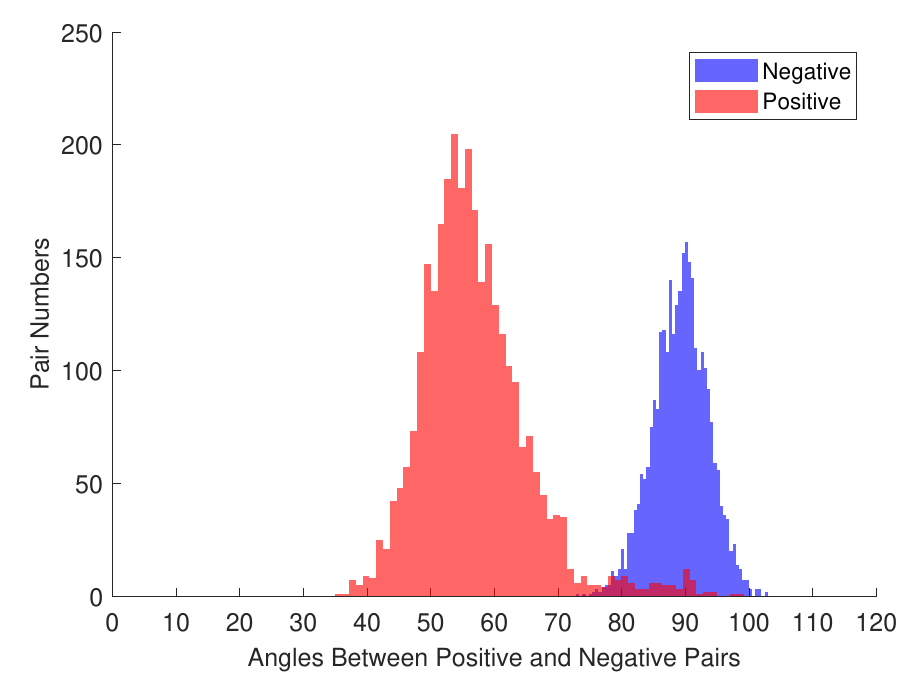}}
\subfigure[CALFW ($96.10\%$)]{
\label{fig:CALFWhist}
\includegraphics[width=0.23\textwidth]{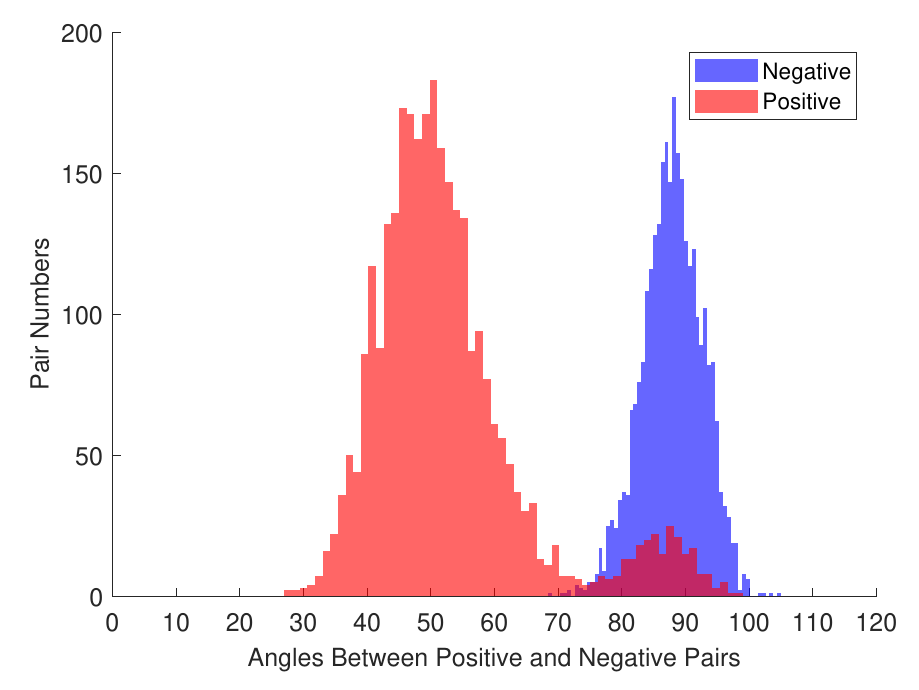}}
\caption{Angle distributions of both positive and negative pairs on LFW, YTF, CFP-FP, CPLFW, AgeDB and CALFW. The red histogram indicates positive pairs while the blue histogram indicates negative pairs. All angles are represented in degree. ([IBUG-500K, ResNet100, ArcFace])}
\vspace{-4mm}
\label{fig:histgramsall}
\end{figure}

\noindent {\bf Results on LFW, YTF, CFP-FP, CPLFW, AgeDB, CALFW.}
LFW \cite{huang2007labeled} and YTF \cite{wolf2011face} datasets are the most widely used benchmark for unconstrained face verification on images and videos. In this paper, we follow the \textit{unrestricted with labelled outside data} protocol to report the performance. As reported in Table \ref{table:lfwytf}, \textcolor{black}{ArcFace models trained on MS1MV3 and IBUG-500K with ResNet100} beat the baselines (\eg SphereFace \cite{liu2017sphereface} and CosFace \cite{tencent2017CosineFace}) on both LFW and YTF, which shows that the additive angular margin penalty can notably enhance the discriminative power of deeply learned features, demonstrating the effectiveness of ArcFace. As the margin-based softmax loss has been widely used in recent methods, the performance begins to be saturated around $99.8\%$ and $98.0\%$ on LFW and YTF, respectively. However, the proposed ArcFace is still among the most competitive face recognition methods.

Besides on LFW and YTF datasets, we also report the performance of ArcFace on the recently introduced datasets (\eg CFP-FP \cite{sengupta2016frontal}, CPLFW \cite{zheng2018cross}, AgeDB \cite{Moschoglou2017AgeDB} and CALFW \cite{zheng2017cross}) which show large pose and age variations. Among all of the recent face recognition models, \textcolor{black}{our ArcFace models trained on MS1MV3 and IBUG-500K} are evaluated as the top-ranked face recognition models as shown in Table \ref{table:calfwandcplfw}, outperforming counterparts by an obvious margin on the pose-invariant and age-invariant face recognition. In Figure \ref{fig:histgramsall}, we show the results of ArcFace model trained on IBUG-500K by illustrating the angle distributions of both positive and negative pairs on LFW, YTF, CFP-FP, CPLFW, AgeDB and CALFW. We can clearly find that the intra-variance due to pose and age gaps significantly increases the angles between positive pairs thus making the best threshold for face verification increasing and generating more confusion regions on the histogram. 

\begin{table}[t!]
\begin{center}
\caption{Face identification and verification evaluation of different methods on MegaFace Challenge1 using FaceScrub as the probe set. ``Id'' refers to the rank-1 face identification accuracy with 1M distractors, and ``Ver'' refers to the face verification TPR at $10^{-6}$ FPR. \textcolor{black}{``R'' refers to data refinement on both probe set and 1M distractors of MegaFace.} ArcFace obtains state-of-the-art performance under both small and large protocols.}
\vspace{-4mm}
\label{table:megaface}
\small
\begin{tabular}{c|c|c}
\hline
Methods  & Id ($\%$) & Ver ($\%$) \\
\hline
Softmax  \cite{liu2017sphereface}                      & 54.85 & 65.92 \\
Contrastive Loss\cite{liu2017sphereface,sun2014deep}   & 65.21 & 78.86\\
Triplet \cite{liu2017sphereface,schroff2015facenet}    & 64.79 & 78.32\\
Center Loss\cite{wen2016discriminative}                & 65.49 & 80.14 \\
SphereFace \cite{liu2017sphereface} & 72.729 & 85.561  \\
CosFace \cite{tencent2017CosineFace} & 77.11 & 89.88 \\
AM-Softmax \cite{wang2018additive}   & 72.47 & 84.44 \\
SphereFace+ \cite{liu2018learning}   & 73.03 & - \\
RegularFace \cite{zhao2019regularface} & 70.23 & 84.07 \\
\hline
CASIA, R50, ArcFace                  & 77.42 & 91.69\\
CASIA, R50, ArcFace, R               & 91.12 & 93.56\\
\hline\hline
FaceNet \cite{schroff2015facenet}      & 70.49       & 86.47 \\
CosFace \cite{tencent2017CosineFace}   & 82.72       & 96.65 \\
UniformFace \cite{duan2019uniformface} & 79.98       & 95.36 \\
RegularFace \cite{zhao2019regularface} & 75.61 & 91.13 \\
AdaptiveFace, R \cite{liu2019adaptiveface} & 95.02       & 95.61 \\
MV-Softmax, R \cite{wang2019mis}      & 98.00  & 98.31 \\
P2SGrad,R \cite{zhang2019p2sgrad}       & 97.25 & -\\
Adocos, R \cite{zhang2019adacos}         & 97.41& -\\
PFE \cite{shi2019probabilistic} & 78.95 & 92.51\\
Fair Loss \cite{liu2019fair} &77.45 & 92.87\\
Search-Softmax, R \cite{wang2020loss} & 96.97  & 97.84 \\
Domain Balancing, R \cite{cao2020domain} & 96.35 &96.56 \\
URFace \cite{shi2020towards}  & 78.60 & 95.04\\
DUL, R \cite{chang2020data} & 98.60 &- \\
CircleLoss, R \cite{sun2020circle} & 98.50 & 98.73\\
CurricularFace, R \cite{huang2020curricularface}  & 98.25  & 98.44 \\
GroupFace, R \cite {kim2020groupface} &  98.74 & 98.79\\
MC-FaceGraph, R \cite{zhang2020global} & {\bf 99.02} & 98.94 \\
SST, R \cite{du2020semi} & 96.27 & 96.96 \\
BroadFace, R \cite{kim2020broadface} & 98.70 & 98.95 \\
\hline
MS1MV3, R100, ArcFace               & 80.71       & 97.46 \\
MS1MV3, R100, ArcFace, R            & 98.51       & 98.74 \\
IBUG-500K, R100, ArcFace          & 81.43       & 97.63  \\
IBUG-500K, R100, ArcFace,R        & 98.98       & {\bf 99.08} \\
\hline
\end{tabular}
\end{center}
\vspace{-4mm}
\end{table}

\noindent {\bf Results on MegaFace.} \textcolor{black}{The MegaFace dataset \cite{kemelmacher2016megaface} includes 1M images of 690K different individuals as the gallery set and 100K photos of $530$ unique individuals from FaceScrub \cite{ng2014data} as the probe set. As we observed an obvious performance gap between identification and verification in the previous work (\eg CosFace \cite{tencent2017CosineFace}), we performed a thorough manual check in the whole MegaFace dataset and found many face images with wrong labels, which significantly affects the performance. Therefore, we manually refined the whole MegaFace dataset and report the correct performance of ArcFace on MegaFace. In Table \ref{table:megaface}, we use ``R'' to denote the refined version of MegaFace and the performance comparisons also focus on the refined version.}

On MegaFace, there are two testing scenarios (identification and verification) under two protocols (large or small training set). The training set is defined as large if it contains more than 0.5M images. For the fair comparison, we train ArcFace on CASIA and IBUG-500K under the small protocol and large protocol, respectively. In Table \ref{table:megaface}, ArcFace trained on CASIA achieves the best single-model identification and verification performance, not only surpassing the strong baselines (\eg SphereFace \cite{liu2017sphereface} and CosFace \cite{tencent2017CosineFace}) but also outperforming other published methods \cite{wen2016discriminative,liu2018learning}.

\begin{figure}[t!]
\centering
\subfigure[CMC]{
\label{fig:megafacecmc}
\includegraphics[width=0.23\textwidth]{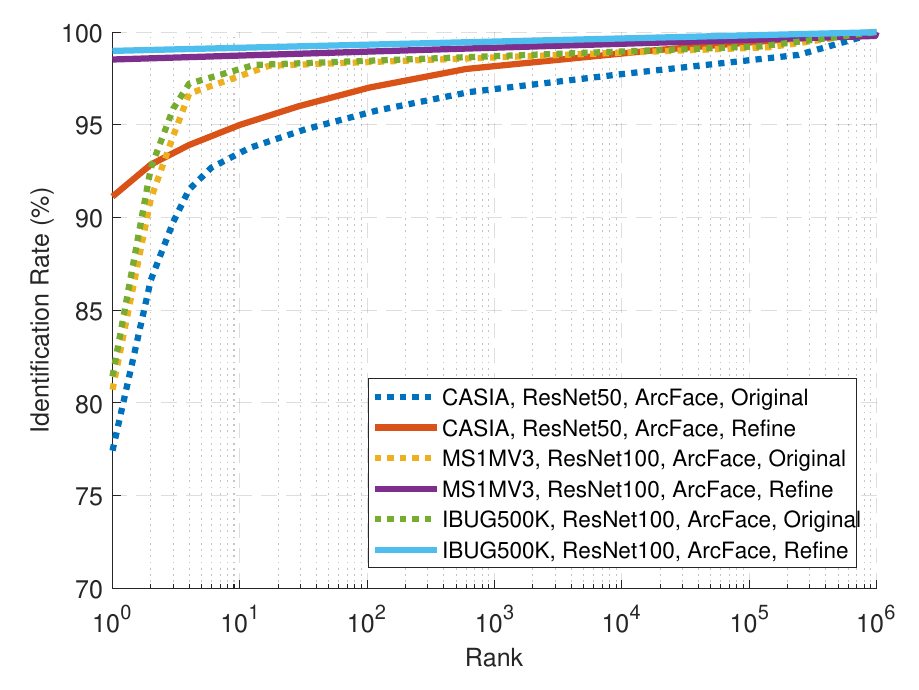}}
\subfigure[ROC]{
\label{fig:megafaceroc}
\includegraphics[width=0.23\textwidth]{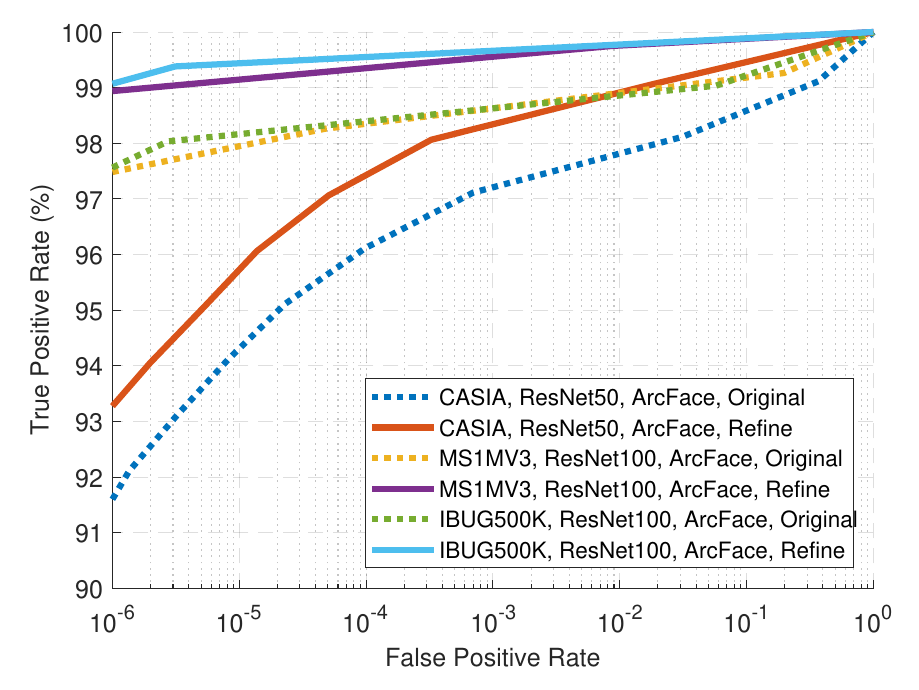}}
\caption{CMC and ROC curves of different models on MegaFace. Results are evaluated on both original and refined MegaFace dataset.}
\vspace{-4mm}
\label{fig:megafacecmcroc}
\end{figure}

\textcolor{black}{
Under the large protocol, ArcFace trained on IBUG-500K surpasses ArcFace trained on MS1MV3 by a clear margin ($0.47\%$ improvement on identification), which indicates that large-scale training data is very beneficial and the proposed sub-center ArcFace is effective for automatic data cleaning under different data scales. As shown in Figure \ref{fig:megafacecmcroc}, ArcFace trained on IBUG-500K forms an upper envelope of other models under both identification and verification scenarios. Compared to MC-FaceGraph \cite{zhang2020global}, ArcFace trained on IBUG-500K obtains comparable results on identification and better results on verification. Considering 18.8M images of 636K identities are used in MC-FaceGraph \cite{zhang2020global}, the performance of our method is very impressive, as we only use images automatically cleaned from noisy web data. Similar to LFW, the identification results on MegaFace are also saturated (around $99\%$). Therefore, the performance gap of $0.04\%$ on identification is negligible and our model is among the most competitive face recognition methods.}

\noindent {\bf Results on IJB-B and IJB-C.} The IJB-B dataset \cite{whitelam2017iarpa} contains $1,845$ subjects with $21.8$K still images and $55$K frames from $7,011$ videos. The IJB-C dataset \cite{whitelam2017iarpa} is a further extension of IJB-B, having $3,531$ subjects with $31.3$K still images and $117.5$K frames from $11,779$ videos. On IJB-B and IJB-C datasets, there are two evaluation protocols, 1:1 verification and 1:N identification.

\begin{table}[t!]
\begin{center}
\caption{1:1 verification (TPR@FPR=1e-4) on IJB-B and IJB-C.}
\vspace{-4mm}
\label{tab:ijb}
\small
\begin{tabular}{c|c|c}
\hline
Method           & IJB-B ($\%$) & IJB-C ($\%$)  \\ 
\hline
ResNet50 \cite{cao2017vggface2} &78.4 & 82.5\\
SENet50 \cite{cao2017vggface2}  &80.0 & 84.0\\
MN-vc \cite{xie2018multicolumn}&83.1 & 86.2\\
DCN  \cite{xie2018comparator}  &84.9 & 88.5\\
Crystal Loss \cite{ranjan2018crystal} & - & 92.29\\
AIM \cite{zhao2019look} & -& 89.5\\
P2SGrad \cite{zhang2019p2sgrad} & - & 92.25\\
Adocos~\cite{zhang2019adacos}   & - & 92.4\\
PFE \cite{shi2019probabilistic} & - & 93.3 \\
MV-Softmax \cite{wang2019mis}   & 93.6 & 95.2 \\
AFRN \cite{kang2019attentional} & 88.5 & 93.1\\
PFE \cite{shi2019probabilistic} &-& 93.25 \\
DUL \cite{chang2020data} &-& 94.61\\
URFace \cite{shi2020towards} &-&96.6\\
CircleLoss \cite{sun2020circle} &-& 93.95\\
CurricularFace \cite{huang2020curricularface} & 94.86 & 96.15 \\
GroupFace \cite {kim2020groupface} & 94.93 & 96.26 \\
BroadFace \cite{kim2020broadface} &94.61& 96.03\\
\hline
VGG2, R50,  ArcFace  & 89.8 & 92.79\\
MS1MV3, R100, ArcFace & 95.42 & 96.83 \\
IBUG-500K, R100, ArcFace & {\bf 96.02} & {\bf  97.27}\\
\hline
\end{tabular}
\end{center}
\vspace{-4mm}
\end{table}

\begin{figure}
\centering
\subfigure[ROC for IJB-B]{
\label{pic:ijbb_roc} 
\includegraphics[width=0.23\textwidth]{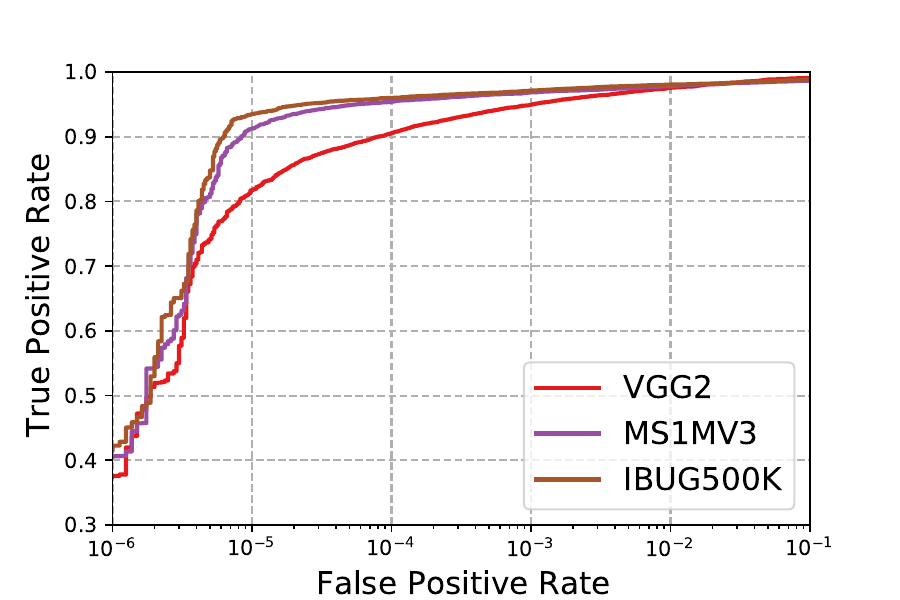}
}
\subfigure[ROC for IJB-C]{
\label{pic:ijbc_roc} 
\includegraphics[width=0.23\textwidth]{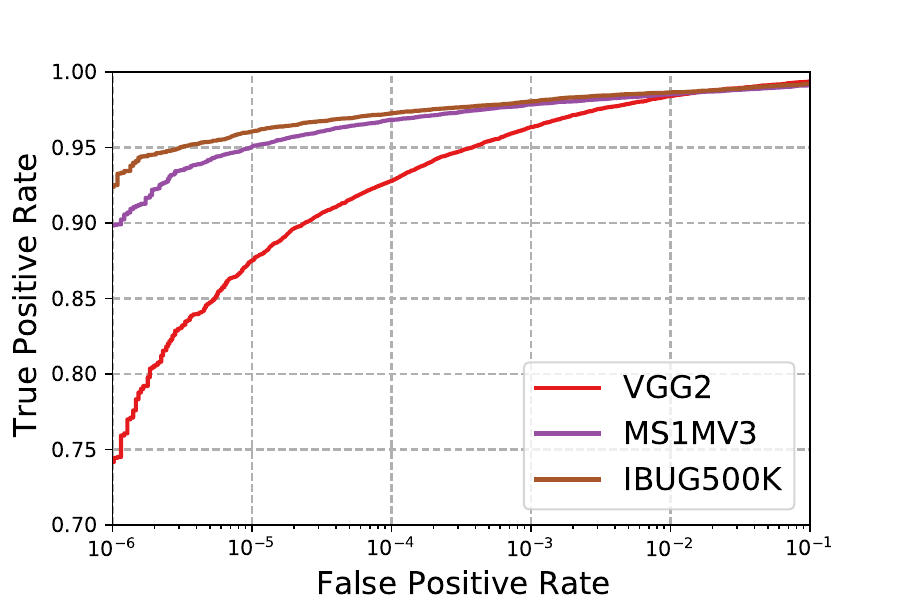}
}
\caption{ROC curves of 1:1 verification protocol on IJB-B and IJB-C. ([Dataset*, ResNet100, ArcFace])}
\vspace{-4mm}
\label{pic:ijb}
\end{figure}

\begin{table}
\begin{center}
\caption{1:1 verification (TPR@FPR=1e-5) and 1:N identification (Rank-1) on IJB-B and IJB-C. ([Dataset*, ResNet100, ArcFace])}
\vspace{-4mm}
\label{tab:ijbdatasetchanges}
\small
\begin{tabular}{c|cc|cc}
\hline
\multirow{2}{*}{Training Datasets } &  \multicolumn{2}{c}{IJB-B} & \multicolumn{2}{c}{IJB-C}\\ 
                         \cmidrule(r){2-3} \cmidrule{4-5}
                        &Ver.($\%$) & Id.($\%$) &Ver.($\%$) & Id.($\%$) \\
\hline
CASIA \cite{yi2014learning} & 62.42 & 86.70& 69.61 & 88.05 \\
IMDB-Face \cite{wang2018devil} & 64.87 & 93.41&66.85& 94.52 \\
VGG2 \cite{cao2017vggface2} & 41.64 & 93.20 & 59.33 & 94.44 \\
MS1MV1 \cite{deng2017marginal} &80.27 & 92.19 &88.16 &93.54 \\
MS1MV2 \cite{deng2019arcface} &89.33 & 94.50 &93.15 & 95.72 \\
MC-FaceGraph \cite{zhang2020global} & 92.82 & 95.76 & 95.62 & 96.93 \\
\hline
MS1MV3  & 91.27  & 95.04 & 95.56 & 96.94 \\
IBUG-500K & {\bf 93.48} & {\bf 95.94} & {\bf 96.07} & {\bf 97.21} \\
\hline
\end{tabular}
\end{center}
\vspace{-4mm}
\end{table}

For the widely used 1:1 verification protocol, there are $12,115$ templates with $10,270$ genuine matches and $8$M impostor matches on IJB-B, and there are $23,124$ templates with $19,557$ genuine matches and $15,639$K impostor matches on IJB-C. In Table \ref{tab:ijb}, we compare the TPR (@FPR=1e-4) of ArcFace with the previous state-of-the-art models. We first employ the VGG2~\cite{cao2017vggface2} dataset as the training data and the ResNet50 as the embedding network to train ArcFace for the fair comparison with the most recent softmax-based methods \cite{cao2017vggface2,xie2018multicolumn,xie2018comparator}. As we can see from the results, the proposed additive angular margin can obviously boost the performance on both IJB-B and IJB-C compared to the softmax loss (about $3\sim5\%$, which is a significant reduction in the error). 

Drawing support from more training data (IBUG-500K) and deeper neural network (ResNet100), ArcFace can further improve the TPR (@FPR=1e-4) to $96.02\%$ and $97.27\%$ on IJB-B and IJB-C, respectively. Compared to the joint margin-based and mining-based method (\eg CurricularFace \cite{huang2020curricularface}), our method further decreases the error rate by $22.57\%$ and $29.09\%$ on IJB-B and IJB-C, which indicates that the automatically cleaned data by the proposed sub-center ArcFace are effective to boost the performance. In Table \ref{tab:ijbdatasetchanges}, we compare the proposed sub-center ArcFace with FaceGraph \cite{zhang2020global} on large-scale cleansing. In FaceGraph \cite{zhang2020global}, one million celebrities (87.0M face images) \cite{guo2016ms} are cleaned into a noise-free dataset named MC-FaceGraph (including 18.8M face images of 636.2K identities) by employing a global-local graph convolutional network. Even though the proposed sub-center ArcFace is only applied to half million identities, the cleaned dataset, IBUG-500K (including 11.96M face images of 493K identities), still outperforms MC-FaceGraph \cite{zhang2020global}. Under the evaluation metric of TPR@FPR=1e-5, the ArcFace model trained on IBUG-500K surpasses the counterpart trained on MC-FaceGraph by $0.66\%$ and $0.45\%$ on IJB-B and IJB-C, respectively. In Figure \ref{pic:ijb}, we show the full ROC curves of the proposed ArcFace on IJB-B and IJB-C, and ArcFace achieves impressive performance even at FPR=1e-6 setting a new baseline. 

For the 1:N end-to-end mixed protocol, there are $10,270$ probe templates containing $60,758$ still images and video frames on IJB-B, and there are $19,593$ probe templates containing $127,152$ still images and video frames on IJB-C. 
In Table \ref{tab:ijbdatasetchanges}, we report the Rank-1 identification accuracy of our method compared to baseline models. ArcFace trained on IBUG-500K achieves impressive performance on both IJB-B ($95.94\%$) and IJB-C ($97.21\%$), setting a new record on this benchmark.

\noindent {\bf Results on LFR2019-Image and LFR2019-Video.} \textcolor{black}{Lightweight Face Recognition (LFR) Challenge \cite{deng2019lightweight} targets on bench-marking face recognition methods under strict computation constraints (\ie computational complexity $<$ 1.0 GFlops). For a fair comparison, all participants in the challenge must use MS1MV3 \cite{deng2019lightweight} as the training data. On LFR2019-Image, trillion-level pairs between gallery and probe set are used for evaluation and TPR@FPR=1e-8 is selected as the main evaluation metric. On LFR2019-Video, billion-level pairs between all videos are used for evaluation and TPR@FPR=1e-4 is employed as the main evaluation metric.}

\begin{table}[t!]
\begin{center}
\vspace{-2mm}
\caption{\textcolor{black}{Verification results ($\%$) on the LFR2019-Image (TPR@FPR=1e-8) and LFR2019-Video (TPR@FPR=1e-4) datasets. ([Dataset*, Network*, ArcFace])}}
\vspace{-4mm}
\label{table:LFR2019}
\small
\begin{tabular}{c|c|c}
\hline
Methods      & Image     & Video  \\
\hline
YMJ$^1$ \cite{zhang2019vargnet} & {\bf 88.78} & -\\
count$^2$ \cite{li2019airface}   & 88.42 & - \\
NothingLC$^3$ & 88.14 & - \\
\hline
NothingLC$^1$ & - &63.23 \\
Rhapsody$^2$  & - &61.87 \\
xfr   $^3$    & - &61.05 \\
\hline
Our Method & 88.65 & {\bf 63.60} \\
\hline\hline
\textcolor{black}{MS1MV3, EfficientNet-B0, ArcFace} & 86.44 & 61.47\\    
MS1MV3, R100, ArcFace         & 92.75 &64.89 \\
\hline
\end{tabular}
\end{center}
\vspace{-4mm}
\end{table}

\begin{figure}
\centering
\subfigure[ROC for LFR2019-Image]{
\label{pic:LFR2019Image} 
\includegraphics[width=0.23\textwidth]{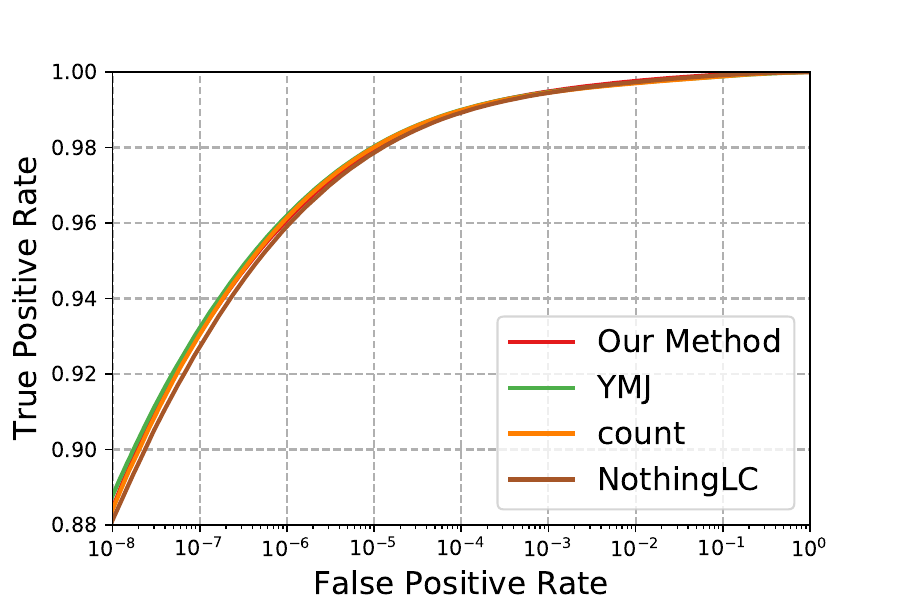}
}
\subfigure[ROC for LFR2019-Video]{
\label{pic:LFR2019Video} 
\includegraphics[width=0.23\textwidth]{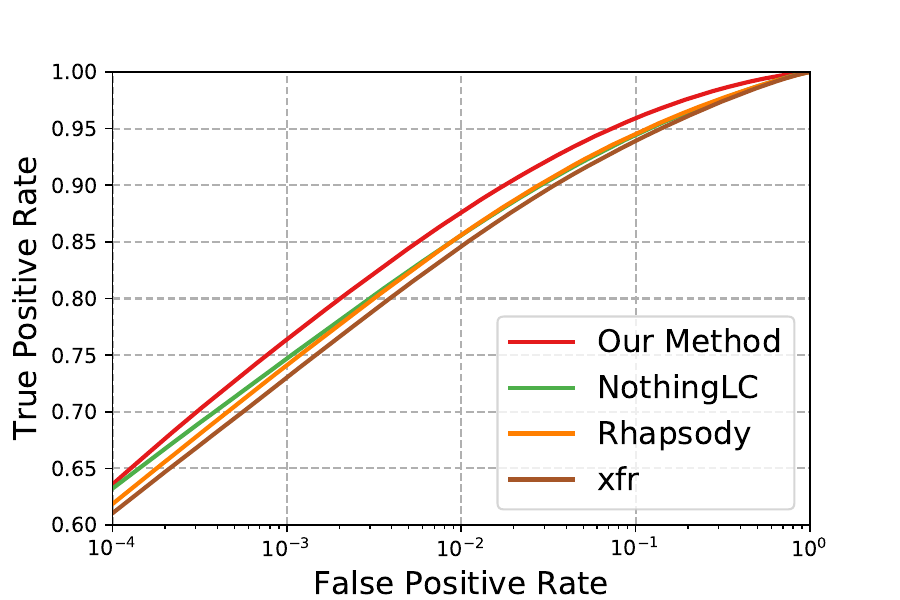}
}
\caption{\textcolor{black}{ROC curves of 1:1 verification protocol on the LFR2019-Image and LFR2019-Video datasets. ([MS1MV3, EfficientNet-B0, ArcFace])}}
\vspace{-4mm}
\label{pic:LFR2019}
\end{figure}

\textcolor{black}{
In Table \ref{table:LFR2019}, we compare the performance of ArcFace with the top-ranked competition solutions \cite{deng2019lightweight}. For the design of our lightweight model, we explore EfficientNet-B0~\cite{tan2019efficientnet} as the backbone. When training from scratch with the proposed ArcFace loss, EfficientNet-B0 can obtain $86.44\%$ on LFR2019-Image and $61.47\%$ on LFR2019-Video, respectively. Following the top-ranked solutions, we also employ knowledge distillation \cite{hinton2015distilling} to boost the performance of our lightweight model. ArcFace trained on MS1MV3 with ResNet100 provides a high-performance teacher network, achieving $92.75\%$ on LFR2019-Image and $64.89\%$ on LFR2019-Video. With the assistance of the teacher network, our lightweight model is trained by minimizing (1) the ArcFace loss (2) the $\ell_2$ regression loss between $512$-D features of the teacher and student networks, and (3) the KL loss \cite{hinton2015distilling} between class-wise similarities predicted by the teacher and student networks. The weights of the $\ell_2$ regression loss and the KL loss is set to $1.0$ and $0.1$, respectively. With knowledge distillation, our method finally achieves $88.65\%$ on LFR2019-Image and $63.60\%$ on LFR2019-Video. As shown in Figure \ref{pic:LFR2019}, our method obtains comparable performance with the champion of the LFR2019-Image track and envelops the ROC curves of all top-ranked challenge solutions in the LFR2019-Video track, surpassing the champion by $0.37\%$.}

\subsection{Inversion of ArcFace}

\begin{figure}
\centering
\includegraphics[width=0.49\textwidth]{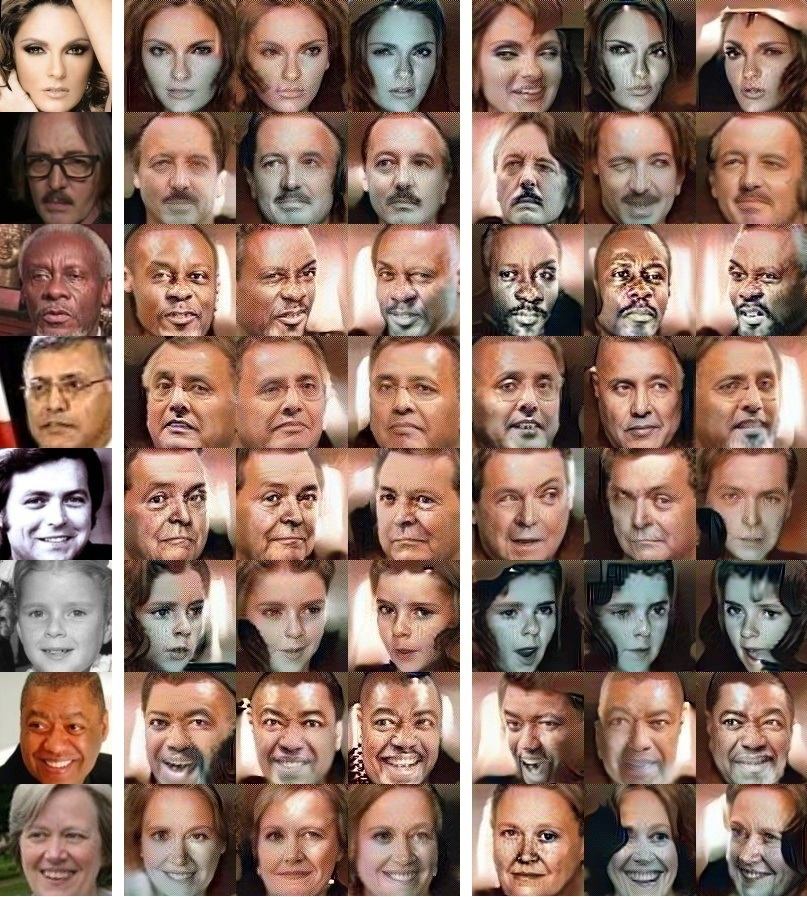}
\caption{Close-set face generation. ArcFace can generate identity-preserved face images only by using the model parameters without training any additional discriminator and generator like in GAN. The first column is the identity from the training data. Column 2 to 4 are the outputs from our ArcFace model. Column 5 to 7 are the outputs from the baseline CosFace model.}
\vspace{-4mm}
\label{fig:closesetgen}
\end{figure}

\begin{table}[t!]
\begin{center}
\caption{FID and cosine similarity of different model inversion results. ArcFace model (ResNet50) for inversion is trained on MS1MV3, but the generated face images also exhibit high similarity from the view of the more powerful ArcFace model (ResNet100) trained on IBUG-500K. The margin parameter for each method is given in the bracket.}
\vspace{-4mm}
\label{table:closesetqualititive}
\small
\begin{tabular}{c|cc}
\hline
Method     & FID        & Cosine Similarity \\
\hline
Softmax    & 75.59    & 0.5612  \\
SphereFace (1.35) & 73.18    & 0.5919 \\
CosFace (0.35)   & 71.64    & 0.6176\\
ArcFace (0.5)    & {\bf 70.39} & {\bf 0.6248} \\  
\hline
\end{tabular}
\end{center}
\vspace{-4mm}
\end{table}

\begin{table}[t!]
\begin{center}
\caption{FID, cosine similarity and verification accuracy on LFW of different model inversion results. The cosine similarity and the verification accuracy are tested by the ArcFace model (ResNet100) trained on IBUG-500K. The margin parameter for each method is given in the bracket.}
\vspace{-4mm}
\label{table:opensetinversequalititive}
\small
\begin{tabular}{c|ccc}
\hline
Method  & FID  &  Cosine Sim & LFW Acc ($\%$) \\
\hline
Softmax    & 77.85    & 0.5504  & 90.14\\
SphereFace (1.35) & 75.16    & 0.5687  & 92.05\\
CosFace (0.35)   & 74.02    & 0.5762  & 92.69\\
ArcFace (0.5)   &\bf{73.16}&\bf{0.5849}  & \bf{93.30}\\ 
\hline
\end{tabular}
\end{center}
\vspace{-4mm}
\end{table}

\begin{figure*}
\centering
\subfigure[ArcFace Inversion for the Young]{
\label{fig:opensetyoung}
\includegraphics[width=1.0\textwidth]{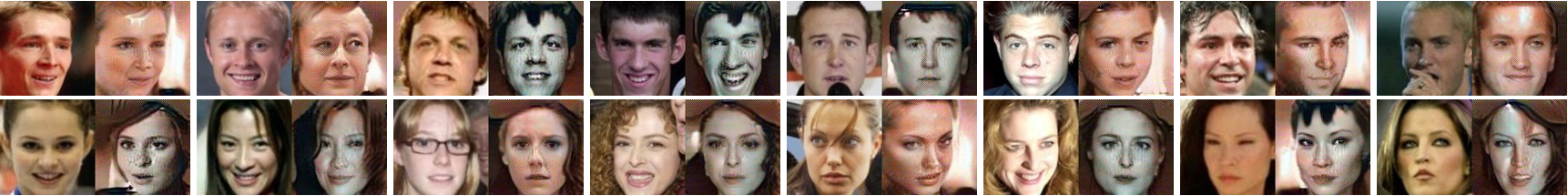}}
\subfigure[ArcFace Inversion for the Old]{
\label{fig:opensetold}
\includegraphics[width=1.0\textwidth]{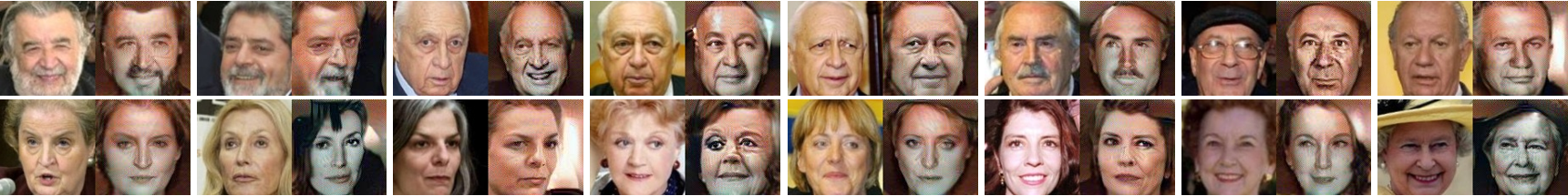}}
\subfigure[ArcFace Inversion for Different Races]{
\label{fig:opensetrace}
\includegraphics[width=1.0\textwidth]{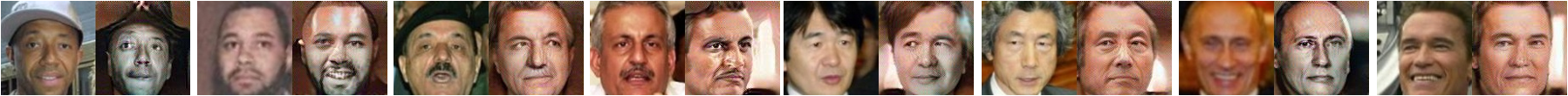}}
\subfigure[ArcFace Inversion under Pose Variations]{
\label{fig:opensetpose}
\includegraphics[width=1.0\textwidth]{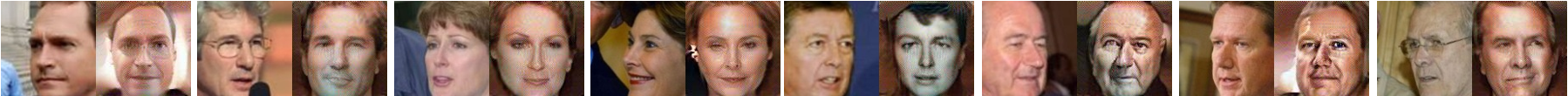}}
\subfigure[ArcFace Inversion under Occlusions]{
\label{fig:opensetocclusion}
\includegraphics[width=1.0\textwidth]{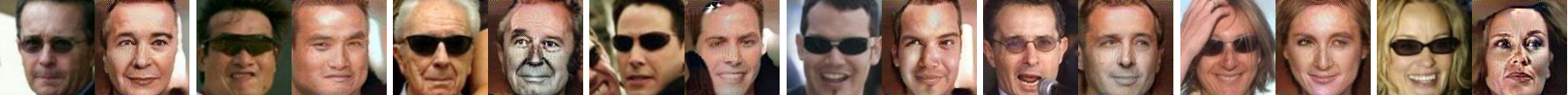}}
\subfigure[Bad Cases of ArcFace Inversion (Gender Confusion)]{
\label{fig:opensetbadcases}
\includegraphics[width=1.0\textwidth]{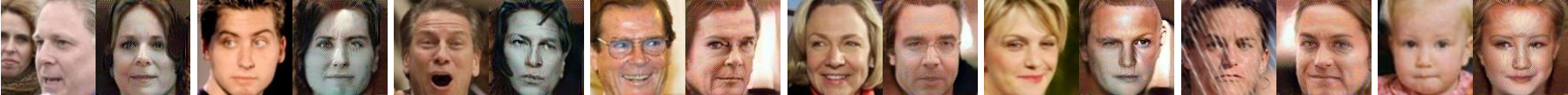}}
\caption{Open-set face generation from the pre-trained ArcFace model. We show the ArcFace inversion results (right) under age, gender, race, pose and occlusion variations by only using the embedding features from LFW \cite{huang2007labeled} test samples (left). In the bottom, we show some bad cases (\eg gender confusion) generated from the ArcFace inversion.}
\vspace{-4mm}
\label{fig:opensetinverselfw}
\end{figure*}

\begin{figure*}
\centering
\includegraphics[width=1.0\textwidth]{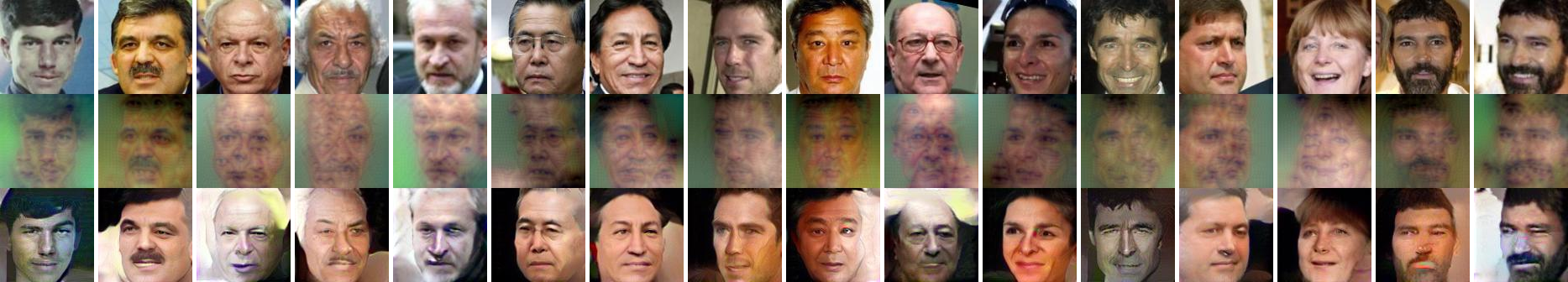}
\caption{Open-set face generation without and with BN constraints. The first row is the original LFW \cite{huang2007labeled} samples. The second row is the ArcFace inversion results without BN constraints, and the third row is the ArcFace inversion results with BN constraints.}
\vspace{-4mm}
\label{fig:opensetinverselfwBN}
\end{figure*}

This section demonstrates the capability of the proposed ArcFace model in terms of effectively synthesizing identity-preserved face images from subject's centers (the close-set setting) or features (the open-set setting).

We adopt the ArcFace (ResNet50) trained on MS1MV3 to conduct the inversion experiments, which include two settings, \ie close-set and open-set. \textcolor{black}{In the close-set mode, centers stored in the linear layer are selected as the targets to generate face images. Identity preservation is constrained by a classification loss (\eg Softmax, SphereFace, CosFace and ArcFace).
In the open-set mode, embedding features predicted by the pre-trained models are used as the targets to generate face images. Identity preservation is constrained by a $\ell_2$ loss. For each time, we synthesize $256$ face images of different identities at the resolution of $112\times112$ in one mini-batch using one NVIDIA V100 GPU.} We employ Adam optimizer \cite{kingma2014adam} at a learning rate of $0.25$ and the iteration lasts 20K steps. Regularization parameters \cite{mordvintsev2015inceptionism} for total variance and $\ell_2$ norm of the generated faces are set as $1e-3$ and $1e-4$, respectively.

In order to quantitatively validate how well the proposed method can preserve the identity of the subject and how \textcolor{black}{visually plausible} the reconstructed face image is, three metrics are adopted: (1) Frechet Inception Distance (FID) \cite{heusel2017gans}; (2) cosine similarity from a third-party model ([IBUG-500K, ResNet100, ArcFace]); and (3) face verification accuracy on LFW for open-set experiments.

\noindent{\bf Close-set Face Generation.} 
In Table \ref{table:closesetqualititive}, we quantify the realism and identity preservation of the reconstructed faces from different face recognition models. For each model, we synthesize training identities by using the 5K randomly selected class indexes. \textcolor{black}{For each identity, different random inputs are gradually updated by the network gradient into identity-preserved face images.} The proposed ArcFace model obviously outperforms the baseline methods (\eg softmax, SphereFace and CosFace) in the image quality, achieving the FID score of $70.39$. By employing the powerful ArcFace model trained on IBUG-500K, we calculate all cosine similarities between real training faces and corresponding generated faces. The average cosine similarity of ArcFace is $0.6248$, surpassing all the baseline models by a clear margin.

In Figure \ref{fig:closesetgen}, we show the synthesized faces from the proposed ArcFace in comparison with the baseline CosFace model. As can be seen, ArcFace is able to reconstruct identity-preserved faces only by using the model parameters without training any additional discriminator and generator like in GAN \cite{goodfellow2014generative}. Considering the image quality is only constrained by the classification loss and the BN priors, it is quite understandable that there exist some identity-unrelated artifacts in the generation results. Besides, there are many grey images in MS1MV3 and this statistic information is also stored in the BN parameters, thus some generated faces are not colorful. Compared to the baseline CosFace model, our ArcFace can depict better facial features of the real faces in terms of identity preservation and image quality. 

\noindent{\bf Open-set Face Generation.} 
In Table \ref{table:opensetinversequalititive}, we compare inversion results of different models on LFW. \textcolor{black}{
For each pre-trained model, we first calculate the embedding features of 13,233 face images from LFW, and then we generate faces constrained to these target features through a $\ell_2$ loss.} As we can see, ArcFace maintains best reconstruction quality and identity preservation, consistently outperforming the baseline models in both FID and average cosine similarity metrics. On the real faces of LFW, the ArcFace model (ResNet50) achieves $99.81\%$ verification accuracy. On the generated faces, the verification accuracy slightly drops to $97.75\%$ by using the same model ([MS1MV3, ResNet50, ArcFace]) for testing. For unbiased evaluation, we report the matching accuracy on LFW by employing the powerful ArcFace model (ResNet100) trained on IBUG-500K and this model is more susceptible to artifacts in the generated results. Even though there is a further drop in the verification accuracy ($93.30\%$), the results compared to the baseline models further demonstrate the advantages of ArcFace in the inversion problem. 

Figure \ref{fig:opensetinverselfw} illustrates our synthesis from features of LFW faces that contain appearance variations (\eg age, gender, race, pose and occlusion). Similar to the previous experiment, our ArcFace model robustly depicts identity-preserved faces. The success of robustly handling with those challenging factors comes from two properties: (1) the ArcFace network was trained to ignore those facial variations in its embedding features, and (2) real face distributions stored in the BN layers can be effectively exploited for face image synthesis. Even though ArcFace can inverse most of the faces with realism and identity preservation, there exist some confusions during generation. In Figure \ref{fig:opensetbadcases}, we show some inversion results from ArcFace containing gender confusions. Even though these confusions can be easily distinguished by human eyes, they exhibit high similarity from the view of the machine. \textcolor{black}{In Figure \ref{fig:opensetinverselfwBN}, we further conduct an ablation study about ArcFace inversion without BN constraints. As we can see from these results, constraints from the BN layers can enforce the generated face more \textcolor{black}{visually plausible}. Without the BN constraints, the resulting face images lack natural image statistics and can be quite easily identified as unnatural.}

\section{Conclusions}

In this paper, we first propose an Additive Angular Margin Loss function, named ArcFace, which can effectively enhance the discriminative power of deep feature embedding for face recognition. We further introduce sub-class into ArcFace to relax the intra-class constraint under massive real-world noises. The proposed sub-center ArcFace encourages one dominant sub-class that contains the majority of clean faces and non-dominant sub-classes that include hard or noisy faces. This automatic isolation can be employed to clean large-scale web faces and we demonstrate that our method consistently outperforms the state of the art through the most comprehensive experiments. Apart from enhancing discriminative power, ArcFace can also strengthen the model's generative power, mapping feature vectors to face images. The pre-trained ArcFace model can generate identity-preserved face images for both subjects inside and outside the training data only by using the network gradient and BN priors. \textcolor{black}{As the proposed ArcFace inversion only focuses on approximating the target identity feature, the facial poses and expressions are not controllable. In the future, we will explore controlling intermediate neuron activations to target specific facial poses and expressions during inversion. In addition, we will also explore how to make the face recognition model not invertible so that face images cannot be easily reconstructed from model weights to protect privacy.} 

\section*{Acknowledgment}

We are thankful to NVIDIA for the hardware donation and Amazon Web Services for the cloud credits. The work of Jiankang Deng was partially funded by Imperial President's PhD Scholarship. \textcolor{black}{The work of Jing Yang was partially funded by the Vice-Chancellor's PhD Scholarship from University of Nottingham.} The work of Stefanos Zafeiriou was partially funded by the EPSRC Fellowship DEFORM: Large Scale Shape Analysis of Deformable Models of Humans (EP/S010203/1), FACER2VM: Face Matching for Automatic Identity Retrieval, Recognition, Verification and Management (EP/N007743/1), and a Google Faculty Award.



\begin{IEEEbiography}[{\includegraphics[width=1in,height=1.25in,clip,keepaspectratio]{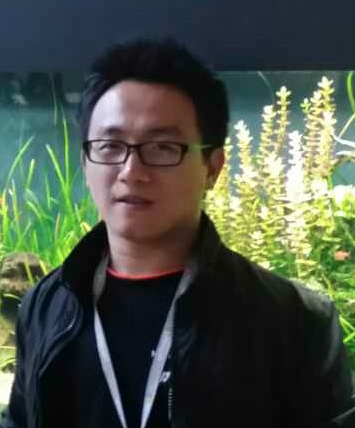}}]{Jiankang Deng} 
obtained his PhD degree from Imperial College London (ICL), supervised by Prof. Stefanos Zafeiriou and funded by the Imperial President's PhD Scholarships. His research topic is deep learning-based face analysis, including detection, alignment, reconstruction, recognition and generation etc. He is a reviewer in prestigious computer vision journals and conferences including T-PAMI, IJCV, CVPR, ICCV and ECCV. He is one of the main contributors to the widely used open-source platform Insightface. He is a student member of the IEEE.
\end{IEEEbiography}

\begin{IEEEbiography}[{\includegraphics[width=1in,height=1.25in,clip,keepaspectratio]{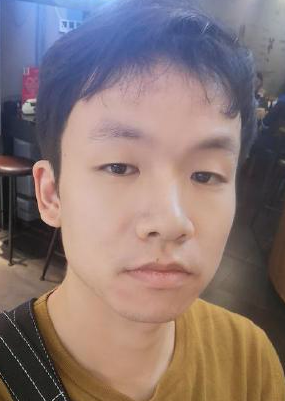}}]{Jia Guo} is an active contributor to the non-profit Github project InsightFace (2D and 3D face analysis).
\end{IEEEbiography}

\begin{IEEEbiography}[{\includegraphics[width=1in,height=1.25in,clip,keepaspectratio]{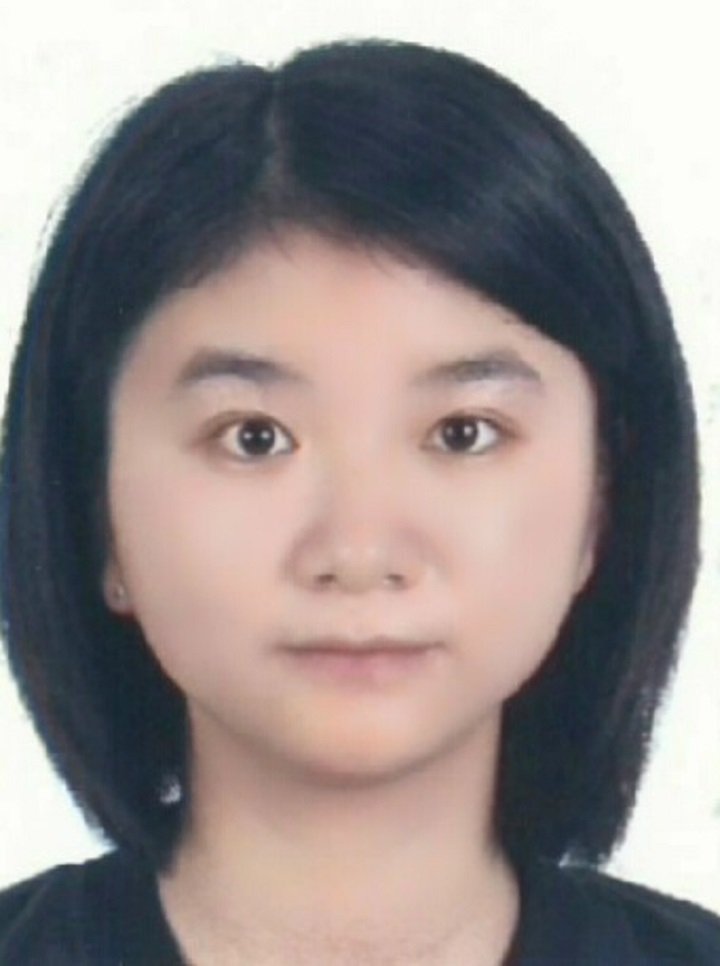}}]{Jing Yang} is a Ph.D. candidate from Department of Computer Science, University of Nottingham. She is funded by the Vice-Chancellor's PhD Scholarship. Her research interest is deep face analysis and model compression.
\end{IEEEbiography}

\begin{IEEEbiography}[{\includegraphics[width=1in,height=1.25in,clip,keepaspectratio]{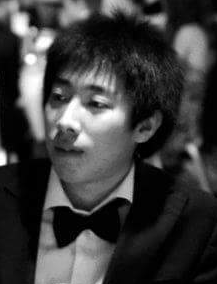}}]{Niannan Xue}
received the BA degree (first class) in theoretical physics from Cambridge University in 2013, and the MMath degree in applied mathematics from Cambridge University in 2014. He is currently working toward the PhD degree at Imperial College London. He was a visiting student in the Biological and Soft Systems Sector of the Cavendish Laboratory. He received St Catharine's Skerne Prize for three consecutive times. His research interests include data mining, machine learning and artificial intelligence. He is a student member of the IEEE.
\end{IEEEbiography}

\begin{IEEEbiography}[{\includegraphics[width=1in,height=1.25in,clip,keepaspectratio]{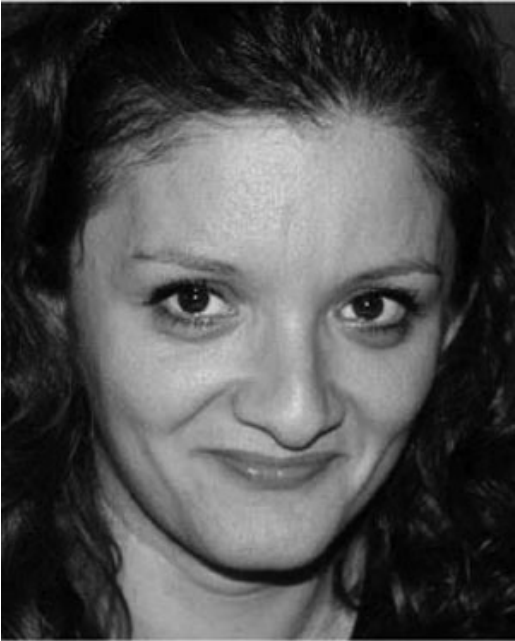}}]{Irene Kotsia}
received the PhD degree from the Department of Informatics, Aristotle University of Thessaloniki, Thessaloniki, Greece, in 2008. 
From 2008 to 2009, she was a research associate and teaching assistant with the Department of Informatics, Aristotle University of Thessaloniki. 
From 2009 to 2011, she was a research associate with the Department of Electronic Engineering and Computer Science, Queen Mary University of London, while from 2012 to 2014, she was a senior research associate with the Department of Computing, Imperial College London. From 2013 to 2020, she was a lecturer in creative technology and digital creativity with the Department of Computing Science, Middlesex University of London. She has been a guest editor of two journal special issues dealing with face analysis topics. She has co-authored more than 40 journal and conference publications in the most prestigious journals and conferences of her field (e.g., the
IEEE Transactions on Image Processing, the IEEE Transactions on Neural Networks and Learning Systems, CVPR, ICCV). She has published one of the most influential works in facial expression recognition in the IEEE Transactions on Image Processing which has received around 700 citations. She is a member of the IEEE.
\end{IEEEbiography}

\begin{IEEEbiography}[{\includegraphics[width=1in,height=1.25in,clip,keepaspectratio]{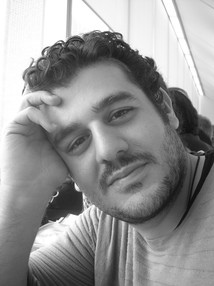}}]{Stefanos Zafeiriou} is currently a Professor in Machine Learning and Computer Vision with the Department of Computing, Imperial College London, London, U.K, and an EPSRC Early Career Research Fellow. He served Associate Editor and Guest Editor in various journals including TPAMI, IJCV, TAC, CVIU,
and IVC. He has been a Guest Editor of 8+ journal special issues and co-organised over 16 workshops/special sessions on specialised computer vision topics in top venues. He has co-authored 70+ journal papers mainly on novel statistical machine learning methodologies applied to computer vision problems, such as 2-D/3-D face analysis, deformable object fitting and tracking, shape from shading, and human behaviour analysis, published in the most prestigious journals in his field of research, such as TPAMI, IJCV, and many papers in top conferences, such as CVPR, ICCV, ECCV, ICML. His students are frequent recipients of very prestigious and highly competitive fellowships, such as the Google Fellowship x2, the Intel Fellowship, and the Qualcomm Fellowship x4. He has more than 20K+ citations to his work, h-index 64. He was the General Chair of BMVC 2017. He is a member of the IEEE.
\end{IEEEbiography}

{
\bibliographystyle{IEEEtran}
\bibliography{egbib}
}

\end{document}